\documentclass[12pt]{article}
\usepackage[utf8]{inputenc}
\usepackage[letterpaper,top=2cm,bottom=2cm,left=2.5cm,right=2.5cm,marginparwidth=1.75cm]{geometry}
\usepackage{graphicx}
\usepackage{url}
\usepackage{csquotes}
\usepackage{breakcites}

\let\oldbibliography\thebibliography
\renewcommand{\thebibliography}[1]{\oldbibliography{#1}
\setlength{\itemsep}{-4pt}}

\title{Reasoning about Procedures with \\Natural Language Processing: A Tutorial}
\author{Li Zhang}
\date{University of Pennsylvania\\\texttt{zharry@seas.upenn.edu}}

\newcommand{\minisection}[1]{\noindent{\bf #1}\hspace{0.6em}}

\begin{document}

\maketitle

\begin{abstract}
    This tutorial provides a comprehensive and in-depth view of the research on procedures, primarily in Natural Language Processing. A procedure is a sequence of steps intended to achieve some goal. Understanding procedures in natural language has a long history, with recent breakthroughs made possible by advances in technology. First, we discuss established approaches to collect procedures, by human annotation or extraction from web resources. Then, we examine different angles from which procedures can be reasoned about, as well as ways to represent them. Finally, we enumerate scenarios where procedural knowledge can be applied to the real world. 
\end{abstract}

\section{Introduction}

Natural language is all about events. This is a not wild claim according to Oxford Languages\footnote{\url{https://languages.oup.com/google-dictionary-en/}}, which defines an event as \textit{a thing that happens}. When people use natural language, it almost always describes something that happens, or events. Events can be grand, like ``glacier movement'', ``establishment of an empire'', or ``world economic crisis''; they can also be tiny, like ``moving one's fingers'', ``breath'', or ``alarm clock ringing.'' All events, great and small, are a fundamental unit of how we describe and perceive the world.

In Natural Language Processing (NLP), the study of events is a trendy field. From 2012 to 2021, the yearly number of papers published on ACL Anthology\footnote{\url{https://aclanthology.org/}} (the primary peer-reviewed paper reserve for NLP) with substring ``event'' in their titles have tripled, from 53 to 154. In comparison, the yearly number of papers with ``translat(e)'' in their titles has remained stable, from 400 to 595. What gives? The importance of events in NLP has been amplified by two factors. First, with the advance of other technologies, such as speech processing, computer vision and computing hardware, more and more real life applications benefit from some knowledge about events. For example, a virtual assistant benefits from knowing the intentions of the user's requests, an advertisement distribution system benefits from modeling user profiles based on users' activities, ans so on. Second, with the advance of NLP tools like large language models, analyzing constructs as abstract as events has become a lot more tractable. Most past NLP work has focused on ``what the texts say'', but now we can attend to ``what is happenning''\cite{chen-etal-2021-event}. This is very exciting, because we have moved one step towards machine understanding of \textit{meaning}. 

\begin{figure}
    \centering
    \includegraphics[width=0.8\textwidth]{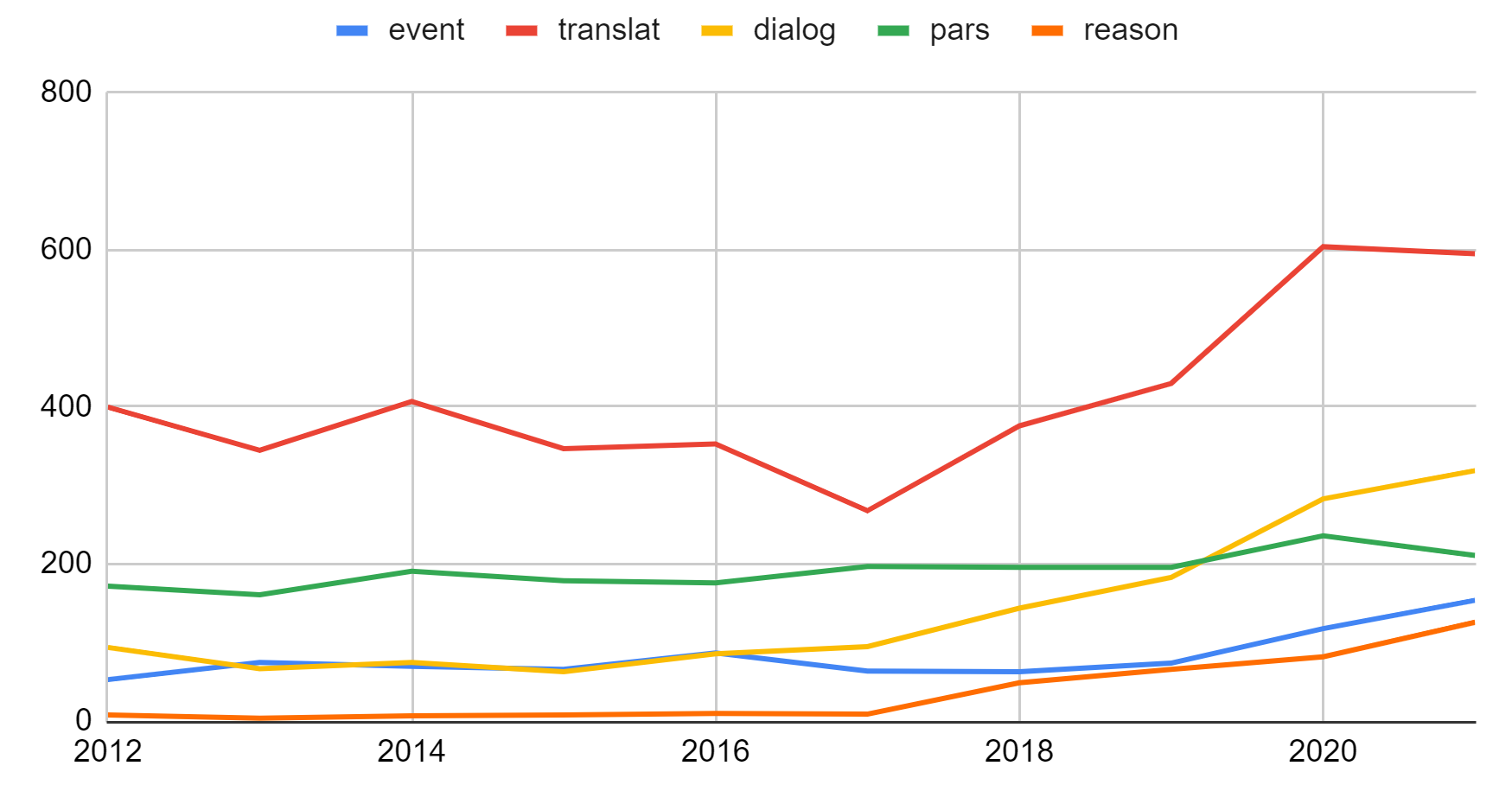}
    \caption{Number of published papers in NLP venues containing keywords of different topics.}
    \label{fig:topic_trend}
\end{figure}

All events are not equal. One special type of event is a \textbf{procedure}\footnote{The term ``procedure'', like many other NLP terms, is overloaded in different disciplines. We follow our own precise definition throughout this tutorial.}. A procedure is a compound event, like ``learn about NLP'', which can be broken down into multiple events, like ``read papers'', ``take classes'', ``attend seminars'', etc. What really distinguishes a procedure from some generic event is that every procedure must consist of some \textbf{goal} (or intent, motivation), and some \textbf{steps} to achieve this goal. As a counterexample, if an individual considers three random things they habitually do in the morning, e.g., ``get out of bed'', ``play tennis'', and ``smoke cigaretts'', these are unlikely to form a meaningful procedure, because a logical goal that these events share is missing. In contrast, ``arrive at the airport'', ``go through securities'', and ``wait at the boarding gate'' are likely part of the procedure ``taking a flight''.  Similarly, events without a sentient perpetrator are unlikely to form a procedure (e.g., ``rain heavily'', ``time elapse``), since they arguably do not have any goal.

Why do researchers care about procedures? The unique roles of events in a procedure, namely, goals and steps, lead to fascinating interactions among the events. The unique challenges and opportunities brought by procedures are manifold compared to ordinary events. For example, a whole class of common sense knowledge is almost exclusive to procedures. Say, a person ``jumps up and down''. There is no telling what they want to achieve. But if this person also ``does push-ups'' and ``stretch arms'', it is then safe to assume that they're ``doing some sort of exercise.'' Such ability to reason about goal not only is interesting in itself with regard to the intelligence of machines, but also has many downstream applications. I should also mention that the NLP community has been so zealous towards common sense reasoning, such that papers with "common" and "sense" in their titles grew from 3 in 2012 to 66 in 2021. However, studying procedures has so much more to offer, as we will see more in this tutorial.

Procedures come in many forms, the most common of which is probably natural language \textbf{instructions}, such as recipes\footnote{Recipes are the most studied procedures in related work by a large margin.}, manual for assembly, how-to guide, navigation lines, etc. An instruction can be seen as a procedure that \textit{should be} carried out instead of one that \textit{is} carried out by someone, just like a recipe provides guidance but is open to interpretation. The majority of past work on procedures has focused on instructions, for good reasons. First, instructions are simple and structured in scope, wordings, presentation, etc., compared to, say, a detailed diary of someone who actually carries out a procedure. Second and more important, instructions can be easily found, on the web, from the books, and so on. In addition, algorithms, programs, execution plans for machines can also be seen as procedures, since they do contain steps for a predetermined goal. More loosely, so are scientific processes, animal behavior, etc. However, let's not attend to them in this tutorial as they are not human activities. Our focus is indeed instructions, but let's keep our sight within procedures in general, since much learning can be shared. 

Can you deal with procedural events the same way you would with general events? The short answer is you could, but you would miss a lot of information. Take the example of the temporal relation among events, which asks the question of which event should happen earlier. For some general events, this can be obvious: Christmas-eve predates Christmas in any calendar year. For procedural events, things get ambiguous. Should you first ``turn off the lights'' and then ``close the door''? That depends on your goal: you should only if you're ``leaving the house'', but not ``entering the house''! It becomes evident that the temporal relation between procedural events is usually conditioned on the goal. Also note that in this example, you physically cannot flip these two events, unless you have a fist-sized hole on your door and you have an arm of several meters. However, this is not always the case. If you're ``meeting with your professor'', you should first ``make an appointment'' and then ``go to the professor's office''. You \textit{could} theoretically flip the steps, but your professor would probably not be happy. In fact, procedural events require specific methodologies in their retrieval, processing, reasoning, and applications, of which we will see more in the tutorial.

TLDR: events are one of the most important things to study in natural language; procedures are one of the most interesting events; instructions are the most common form of procedures. In this tutorial, you will comprehensively learn about procedures from the point of view of NLP. That is, you will first see how past research has tackled procedures, where to find data, how can we reason about them, and what use they have in real life. 

\section{A Brief History of Procedures}
Research on procedures has largely been driven by available techniques and applications of interest. Hence, it is beneficial to review efforts on studying procedures chronologically to understand the change of interest and methodology over time. 

The earliest work on procedures in NLP is probably \cite{miller1976natural}, which is unavailable today. The earliest available work was \cite{momouchi-1980-control}, which aimed to convert a procedure to a comprehensive flowchart. The work formally defined a procedure as ``a sequence of actions (steps)
intended to achieve a goal.'' It primarily focused on \textbf{knowledge acquisition}, which means to extract various features from procedures. A usual byproduct of such extraction is some \textbf{procedural representation}. Many such features are still receiving much attention today, such as pre- and post-conditions, temporal duration and relation, etc. Knowledge acquisition is a heavy focus in this tutorial, as it is the gateway to reasoning about procedures; such knowledge can be applied to myriad of scenarios. 

The procedures studied here were from recipes, a common kind of instructional texts that we will later see used in many works.

Procedural understanding is closely related to \textbf{planning} in artificial intelligence \cite{schank_scripts_1977}. Specifically, a substantial body at that time focused on natural language generation from plans \cite{mellish1989natural, wahlster1993plan}. It was not surprise that these works had found instructional texts tractable \cite{kosseim1994content,paris1995support}, which are structured and limited in scope. However, these works hardly attempted to learn from procedures, until later when \cite{10.1145/584955.584977} created a human-in-the-loop tool for procedural knowledge acquisition. Note that such knowledge extracted from procedural texts can be transformed to plans (the reverse of ``natural language generation from plans'' mentioned above), which is a substantial body of research \cite{macmahon2006walk, branavan-etal-2009-reinforcement,10.5555/2900423.2900560,artzi-zettlemoyer-2013-weakly,kiddon-etal-2015-mise}. We will go into more details later in this tutorial.

One primary use of such procedural language generation is to answer how-to questions, the second most sought-after type of queries on the internet at that time \cite{deRijke2005QuestionAW}. To that end, there were an array of efforts to identify \textbf{instructional texts} on the web \cite{takechi-etal-2003-feature}, automatically generate them \cite{paris_automatically_2005}, converting them to executables \cite{gil2011tellme,fritz2011formal}, study their linguistic idiosyncrasies \cite{https://doi.org/10.1111/0824-7935.00118,bielsa2002semantic,aouladomar2005preliminary,10.1145/2531920}, and extract components such as titles \cite{delpech-saint-dizier-2008-investigating}. 

Procedural knowledge acquisition and representation continued to thrive \cite{lau2009interpreting,Addis2011FromUW}. The next outstanding work on procedural knowledge acquisition would be \cite{zhang-etal-2012-automatically}, which proposed a standard representation of procedures. Compared to \cite{momouchi-1980-control}, the proposed representation also stressed upon goals, steps, pre- and post-conditions, but was much simpler and set the basis of procedural representation in subsequent research. \cite{maeta-etal-2015-framework, kiddon-etal-2015-mise} took a different approach, and focused on the relations among entities in procedures. Note that the entity-relation view has a long history \cite{10.1145/320434.320440} and is a center piece in information extraction in NLP \cite{doddington-etal-2004-automatic,ellis2014overview}. We will see more about representing procedures later in the tutorial. 

Just want kind of procedures are on the web? One most used source is \textbf{wikiHow}\footnote{\url{wikihow.com}} (previously eHow\footnote{\url{https://www.wikihow.com/wikiHow:History-of-wikiHow}}), a website of how-to instructions for many tasks. As the structure and writing style are consistent, wikiHow has been leveraged by NLP researchers since its inception \cite{10.1145/988672.988750,Addis2011FromUW,10.1145/2567948.2578846}. In recent years, wikiHow has grown massively in size (now more than 110k articles), diversity (19 languages, hundreds of categories) and quality (editorial process\footnote{\url{https://www.wikihow.com/wikiHow:Delivering-a-Trustworthy-Experience}}). As a result, it has fueled much research on procedures from various aspects \cite{pareti2018representation}, such as linking actions \cite{pareti2014integrating,10.1145/2872518.2890585,lin-etal-2020-recipe,donatelli-etal-2021-aligning,zhou-etal-2022-show}, what-if reasoning \cite{tandon-etal-2019-wiqa,rajagopal-etal-2020-ask}, entity tracking \cite{tandon-etal-2020-dataset}, next-event prediction \cite{nguyen-etal-2017-sequence,zellers-etal-2019-hellaswag,zhang-etal-2020-analogous}, intent reasoning \cite{dalvi-etal-2019-everything}, goal-step reasoning \cite{zhou-etal-2019-learning-household,park2018learning,zhang-etal-2020-reasoning,yang-etal-2021-visual}, procedure generation \cite{sakaguchi-etal-2021-proscript-partially,lyu-etal-2021-goal}, simulation \cite{puig2018virtualhome}, summarization \cite{DBLP:journals/corr/abs-1810-09305,ladhak-etal-2020-wikilingua} and so on. Those are just some works that use procedures from wikiHow, and there are plenty that use other data sources, that we will examine later. 

Indeed, as more work has explored procedures \cite{9070972}, research agendas become more diverse. With the advance of large language models \cite{devlin-etal-2019-bert,NEURIPS2020_1457c0d6}, researchers no longer fixate on formally representing procedural knowledge, but directly use procedural texts for an array of downstream tasks, as enumerated above. It is truly a great time to study procedures.

\begin{figure}
    \centering
    \includegraphics[scale=0.9]{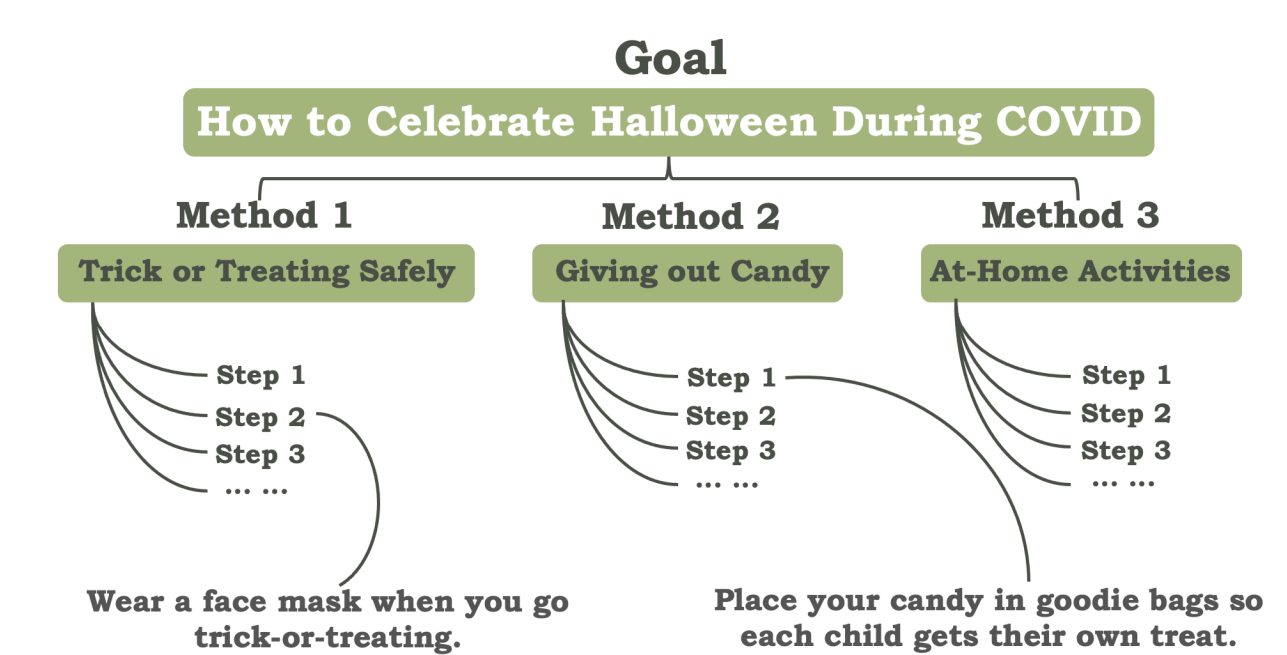}
    \caption{An illustration of an article in wikiHow. Typically, an article contains a goal and some methods, each containing some steps.}
    \label{fig:wikihow}
\end{figure}

\section{Source of Procedures}
Any work that studies procedures must first obtain them. There are two prominent sources: web resources and human annotation. 

\subsection{Web Resources}
Extracting procedures from some resources such as cookbooks has to be one of the most intuitive ways of obtaining procedural data. Nowadays, most such instructions can be found online in the web. As mentioned before, wikiHow is one prominent resource with many general how-to instructions. Let's look at a case study to extract procedural information from wikiHow from one of my own work \cite{zhang-etal-2020-reasoning}.

\minisection{Scrape.} First, we need to scrape the entire wikiHow website using some web-crawler to get the HTML for each wikiHow page. To that end we can use a Python package called Beautiful Soup. To traverse all wikiHow pages, we can use breadth-first search to expand every link found in each page whose URL contains \url{wikihow.com}. After getting the HTML for each wikiHow page, we filter ones corresponding to actual articles using some common characteristics, such as specific formatting of the title, sections, etc.

\begin{figure}
    \centering
    \includegraphics[width=\textwidth]{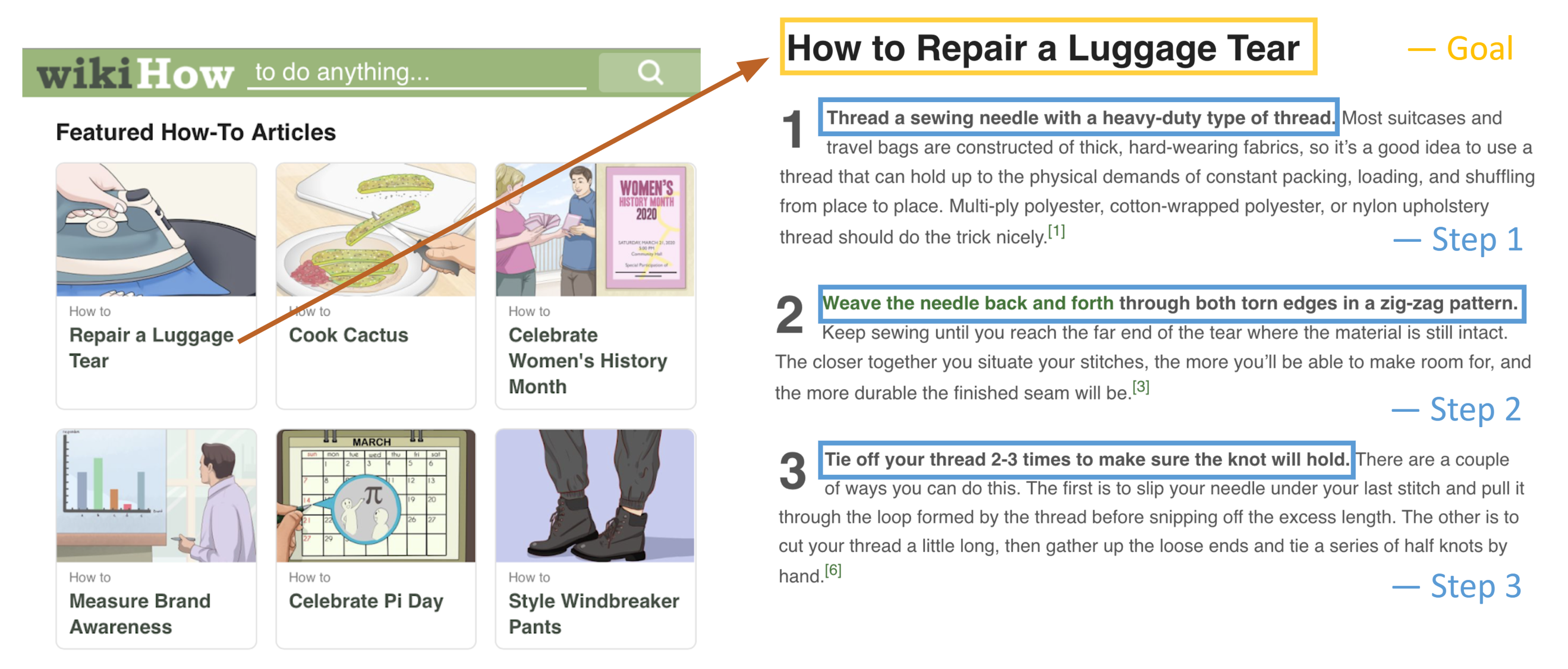}
    \caption{An example of how we can extract a procedrue from a wikiHow article.}
    \label{fig:wikihow_extraction}
\end{figure}

\minisection{Extract.} We must first ascertain what information we would need for a procedure. Minimally, a procedure might have a goal and some steps. The \textit{title} of a wikiHow article naturally suffices as the goal. Alternatively, many wikiHow articles also have \textit{methods} or \textit{parts}, the title of which could also be a more fine-grained goal. Each \textit{step} in a wikiHow article contains:
\begin{itemize}
    \itemsep0em 
    \item a \textit{headline} in bold, which is always the first sentence,
    \itemsep0em 
    \item \textit{details}, which are subsequent sentences,
    \itemsep0em 
    \item \textit{additional information}, which is optionally listed in a bullet points.
    \itemsep0em 
\end{itemize}
See an example in Figure~\ref{fig:wikihow}. One might decide what to include as goal and steps in a procedure depends on their use cases. For now, let's take a minimalist approach and treat the title as the goal, and the headlines as the steps (Figure~\ref{fig:wikihow_extraction}). Using some HTML parser like Beautiful Soup, we can extract them easily. 

The data we have at this point are noisy. For example, some HTML or texts might be malformed. It is thus importantly to manually check some obtained procedures to confirm their quality. Alternatively, you can also use the processed wikiHow procedural data we released\footnote{\url{https://github.com/zharry29/wikihow-goal-step}}. 

\subsection{Human Annotation}
Before instructions became widely available on the web, human annotation has been a viable option to obtain procedural data. Usually, this is done by crowdsourcing, where many people contribute to annotating the data. Compared to using web resources, crowdsourcing costs money and requires careful design. One primary way of getting human annotations is via Amazon Mechanical Turk (mTurk)\footnote{\url{https://www.mturk.com/}}, as exemplified by \cite{lau2009interpreting}. In essence, mTurk is a marketplace where researchers pay people (namely, the crowd) to annotate data. To collect procedures, they ask crowd workers to write down a task and provide its instructions. In their case, they collected a total of 43 procedures with 300 steps, costing only 10 US dollars!\footnote{It is important to pay annotators fairly. By a rough comparison, in 2009 the US minimal wage is \$7.25. To earn it, an annotator needs to annotate 29 procedures per hour, or 1 per 2 minutes.}

With a reasonable budget, crowdsourcing can result in a great deal of annotations. Moreover, it offers much freedom with regard to the type of annotations we want. For example, \cite{regneri-etal-2010-learning,li2012crowdsourcing,wanzare-etal-2016-crowdsourced} each collected data via crowdsourcing with different topics and sizes.

\section{Representation of Procedures}
\label{sec:representation}
At this point we have a collection of procedures, each containing a goal and some steps written in texts. At this simplest form, a procedure is be represented as a tree of depth 1, with the goal as the root and the step as leaves. The edges thus represent ``goal-step'' relations. The events are nodes and represented as plain texts. Such a procedural representation doesn't tell us or the models much, even compared to the ancestral work \cite{momouchi-1980-control} who converted procedures to comprehensive flowcharts. Fortunately, there are much that can be done to represent procedures. 

\begin{figure}[t]
    \centering
    \includegraphics[width=\textwidth]{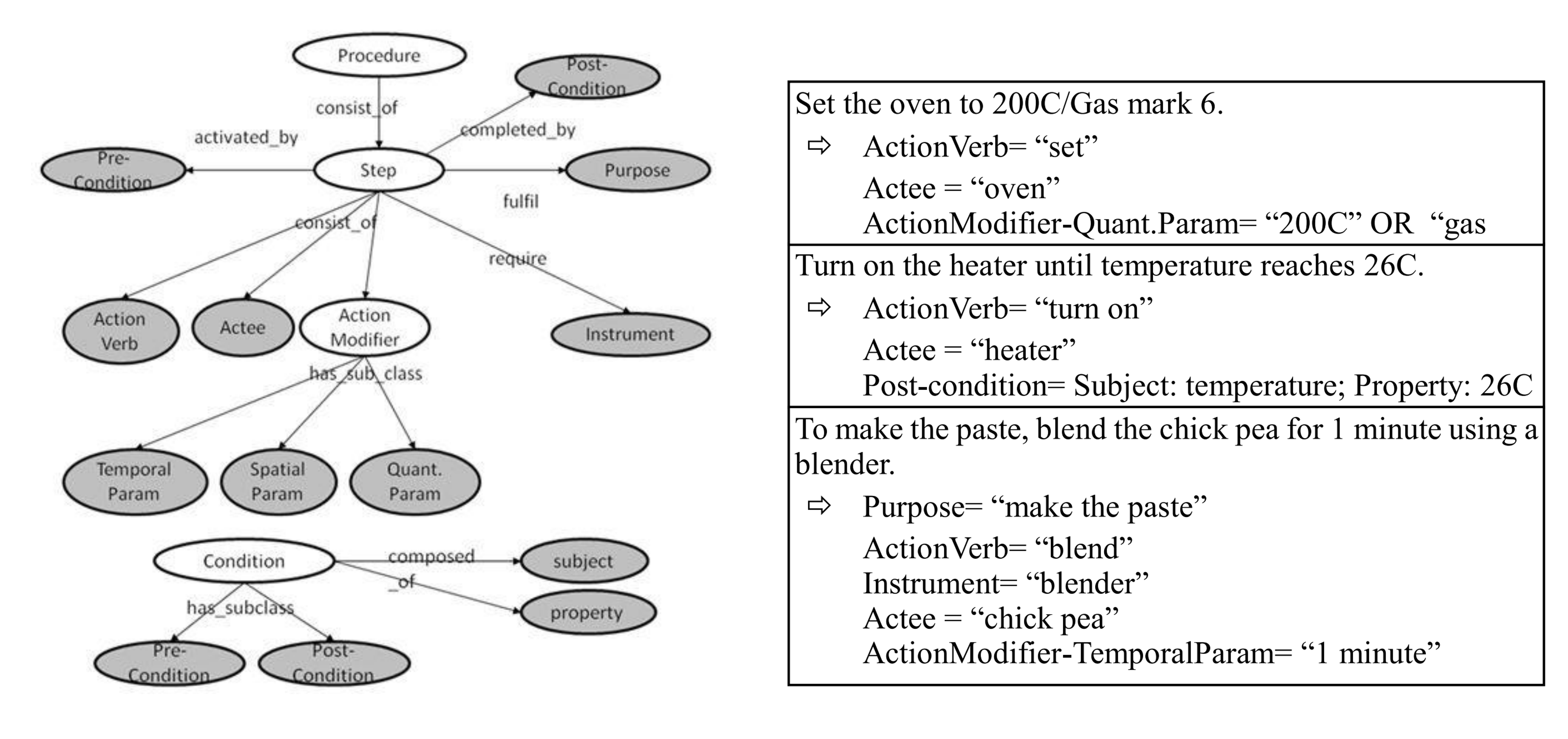}
    \caption{The formal representation of a procedure, and an example instantiating of the components from \cite{zhang-etal-2012-automatically}.}
    \label{fig:zhang2012}
\end{figure}

\subsection{Structure vs. Texts}
Before we dive into more sophisticated representation of procedures, it's worth briefly covering why we want to extract structured knowledge into knowledgebases, databases, or schemata \cite{shapiro2000natural}, and whether doing so is necessary at all. Why aren't we satisfied of representing procedures as natural language sentences? The most obvious reason is that for a long time, machines simply have been better at dealing with symbolic data than textual data. Assume you want to devise a software that can tell you the ingredients of a dish, and where you can buy them. If you are blessed with a knowledgebase with the ingredients of all dishes and the location of all ingredients, you can trivially accomplish the task. Even if some information is missing, you can use machine learning techniques such as K-nearest-neighbors or even neuro-symbolic methods to infer the missing information. However, if all you have is a collection of sentences describing similar information about the ingredients and locations, such as ``I just bought two bottles of milk at a grocery store yesterday,'' machines struggle at leveraging this information without first extracting some structured knowledge. This is strong evidence that structure representation of knowledge, e.g., procedures, is beneficial. No wonder much effort has been put into representing all kinds of knowledge using a unified structure, such as the Linked Data Cloud\footnote{\url{https://lod-cloud.net/}} devised by the semantic web community.

On the flip-side, structured representation can be brittle and inflexible. The knowledgebase can have many potential flaws. It might only cover a subset of information. It might contain partial information\footnote{For example, a knowledgebase can have an entry for ``milk'' but none for ``oat milk,'' and thus cannot deal with the latter at all despite its high similarity to the former.}. It might contain errors. All these issues can easily render machine inference faulty or impossible. 

A surprising contender to structured knowledge representation is to not ``represent'' at all, but instead maintain plain textual representation. In recent years, these natural language sentences can be effectively leveraged by large language models (LLMs) to make effective inferences. We will see more later that LLMs such as GPT-3 \cite{brown2020language} or T5 \cite{2020t5} actually thrives in the void of structure of the data. 

The key takeaway here is that structured representation of procedures is a \textbf{valid, but not the only approach} to reasoning about them. Obviously, the quality of such representation directly influences how well models can do. Let's then check out a couple of ways to represent procedures. 

\subsection{Structured Representation of Procedures}

\cite{zhang-etal-2012-automatically} proposed a reasonable graph representation of procedures. See Figure~\ref{fig:zhang2012} for an illustration. Specifically, a procedure consists of some steps, and of course, a goal. Each step consists of an action verb and an actee. It may also have temporal, spatial or quantitative information. A step is activated by some pre-condition, and completes with some post-condition. A step may also require some instruments, and may fulfill some purpose (sub-goal). Such formulation covers most of the important factors of a procedure. 

Two major problems ensue. First of all, it is challenging to accurate extract all such information from a textual representation of a procedure, such as a recipe from a cookbook. The authors of \cite{zhang-etal-2012-automatically} employ rule-based systems such as parsers to extract the components. However, even were they to use the state-of-the-art neural models as of 2022, the extracted information might not be complete, since some information can be implicit. For example, in the action ``add salt to water and bring to a broil'', knowing ``bring \textit{what} to a broil'' requires co-reference reasoning. Further, features like pre- and post-conditions are even trickier, and remains an open question to this day, that I also actively work on. Without a means to accurately extract such information, it is challenging to apply the representation to downstream tasks such as question answering for a wide variety of procedures.

\begin{figure}[t]
    \centering
    \includegraphics[width=\textwidth]{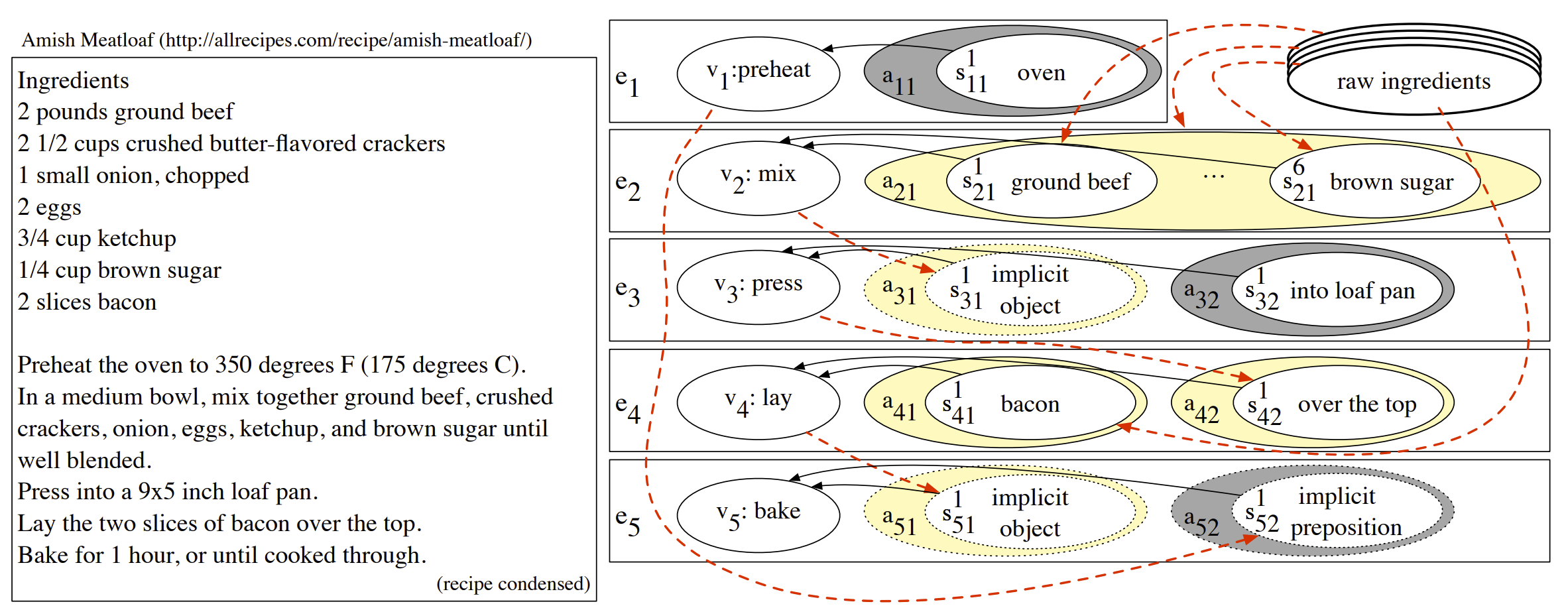}
    \caption{An example recipe and its representation from \cite{kiddon-etal-2015-mise}, denoting the actions, ingredients, and their relations to one another.}
    \label{fig:kiddon2015}
\end{figure}

\cite{kiddon-etal-2015-mise} provides another representation of procedures, focusing on recipes. Compared to the previous representation, their focus shifts to entities and relations in procedures. Naturally, this representation is a boon for answering questions about these entities, such as ``where is the bacon'' or ``what is mixed with the flour''. However, it hinges on the relatively regular phrasings from recipes, and may not handle other types of procedures well. 

\subsection{Textual Representation of Procedures}

Representing procedures as texts is quite trivial, and we have seen how we can achieve this. With the advance of large language models, diverse expressions of procedures in free-form texts can also be directly input into the models. Free from the bottleneck of extracting information to form a structured representation, researchers can harness the power of these language models to learn from procedures and apply learned knowledge to downstream tasks more flexibly. For the recent years, this has been the dominant approach. In the next section, we will go with the flow and see many examples of working with only textual representation. 

Just because language models work excellently with textual representation does not mean that structured representations are in vain. They still remain a formidable force for procedural reasoning, especially coupled with quickly advancing neural-symbolic methods \cite{NEURIPS2021_d367eef1}. Moreover, structured representations are a step towards language grounding, which is crucial to robots executing instructions \cite{puig2018virtualhome,huang2022language,ahn2022can}. This area is of course related to procedures and an extremely important front of artificial intelligence, but we will not go into details in this tutorial. 

\section{Learning from Procedures}
There are many world knowledge encapsulated in procedures, referred to as procedural knowledge. Learning it not only helps with downstream applications, but also sheds light on model's capabilities. In this section, we will see how models can learn some common knowledge from procedures. 

\subsection{Goal-Step Relations}
At the core of a procedures are goals and steps. In practice, however, we might not have the luxury of having all of them at hand. There are many cases where we need to infer the goal from steps. For example, sociologists might be interested in people's motivation of their actions; advertisers might want to know what people want from their browsing activities; dialog systems might need to figure out the intent from utterances about user activities. Conversely, inferring steps from a goal is equally useful. For example, a household assistant might need to help user figure out how to accomplish tasks. Inferring goals or steps hinges on understanding the goal-step relation between events. Moreover, the temporal relation between two steps is equally important. While temporal relations among general events has been well studied \cite{zhou-etal-2019-going,han-etal-2019-joint,vashishtha-etal-2020-temporal,ma-etal-2021-eventplus}, the ordering of steps additionally depends on the goal. For example, ``go to the office'' and ``buy food'' in general do not have a habitual order, but if both events are step from the procedure with the goal of ``have an office potluck'', it becomes clear that ``buy food'' should happen first. Such is an illustrative example of procedural events' idiosyncrasy.

\begin{figure}
    \centering
    \includegraphics[width=\textwidth]{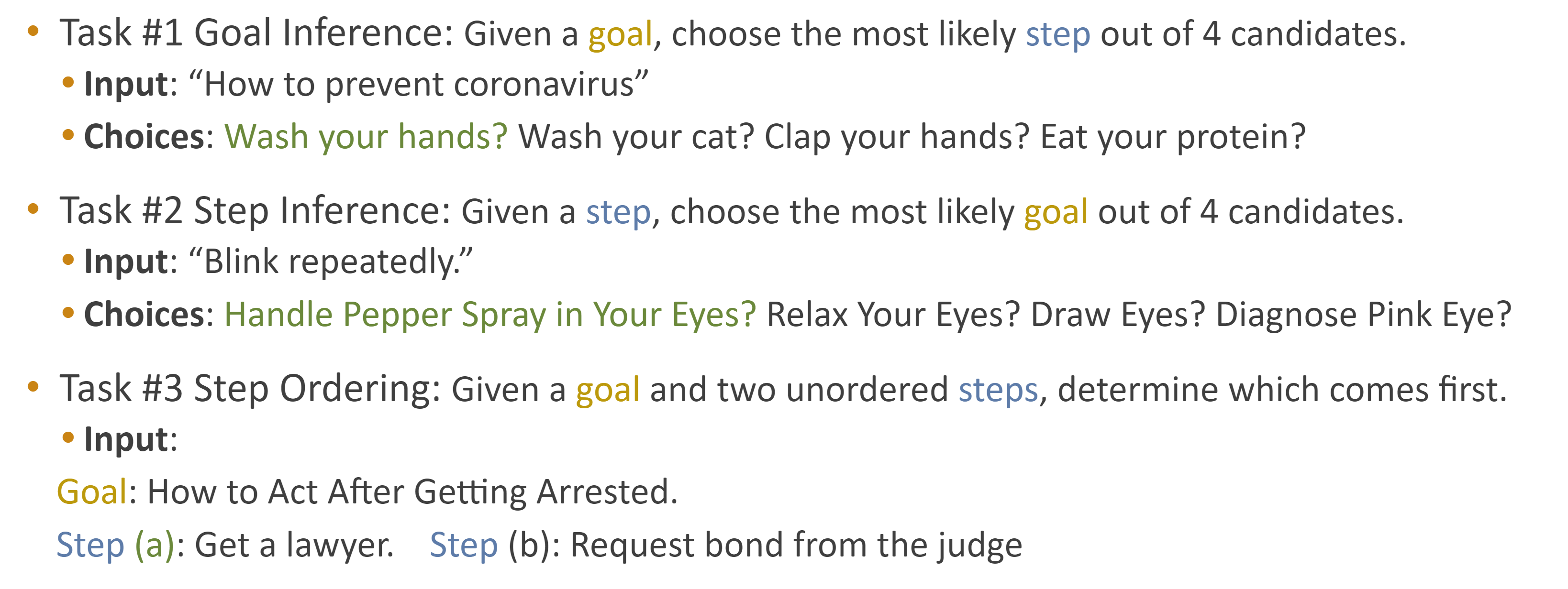}
    \caption{Examples from the goal-step reasoning tasks from \cite{zhang-etal-2020-reasoning}.}
    \label{fig:goal_step}
\end{figure}

Apart from the practical uses, the knowledge of goals and steps is a part of human common-sense and might be a critical part of artificial general intelligence. Being able to reason about goals and steps also has strong implication towards other NLP tasks such as question answering, machine reading comprehension, etc. Let's see how we can learn the goal-step relations in wikiHow following one of our own work \cite{zhang-etal-2020-reasoning} (examples shown in Figure~\ref{fig:goal_step}). 

\subsubsection{Goal-Step Relation} 

The simplest way to formulate this task is that given two events, a model needs to predict whether one can be the step while another can be the goal, logically. In previous sections, we have extracted hundreds of thousands of procedures from wikiHow, each containing a goal and some steps. A pair of such a goal and a step then forms a positive example, while a pair of some goal and a step from another procedure likely forms a negative example. However, directly judging the goal-step relation might be subject to ambiguity. For example, is ``play guitar'' a logical goal of ``stretch your arms''? It really can go both ways, since you realistically can get cramps if you practice guitar for too long and your arms get sore. However, if we consider another potential goal like ``debate'', then the step ``stretch your arms'' is far more unlikely. For this reason, we set up the task comparatively, using a multiple choice setting. Namely, given a goal and a couple of potential steps, a model needs to choose the most likely step. Vice versa for given a step and some goals. Consider this example of inferring the step from a goal.
\begin{displayquote}
What is the most likely step of the goal ``prevent viruses''?\\
A. Wash hands. B. Clap hands. C. Wash a cat. D. Sleep early.
\end{displayquote}
The answer is unambiguously A, as B and C are irrelevant, and while D does help with boosting one's immunisation system, it's less direct than A. 

Now that we formulate the step-inference and goal-inference tasks as multiple choice, an interesting and important consideration is how we should sample negative examples. Note that sampling positive examples is trivial as we can just take a goal and a step from a wikiHow procedure. Think about a multiple choice exam. If the ``wrong options'' are too obvious, the exam becomes trivial. Conversely, all of us must have seen some frustrating ``bad questions'' where there are multiple correct answers, which should not occur. In our case, given a step, how do we sample some non-goals that are distracting enough for the models? One way is to use semantic similarity search, which helps us find sentences that have similar meaning. Let's consider the step ``store your receipts in a binder'' that comes from the wikiHow procedure with the goal ``organize receipt''.  We then need to find a couple of other goals similar to ``organize receipt'', but they cannot be the goal of ``store your receipts in a binder''. If we search for ``organize receipt'' using a search engine on wikiHow, we can find ``write a receipt'', ``print a uber receipt'', ``create a donation receipt'', etc. All those are favorable, since they each cannot be a logical goal of ``store your receipts in a binder''. To implement such a search engine, a typical way is to use Elasticsearch with the classical BM25 search algorithm, while indexing the entire textual information from a wikiHow website. 

An alternative to using search engines is to use a sentence embedding approach. First, we use a sentence encoder such as sBERT  \cite{reimers-gurevych-2019-sentence} to encode each step and each goal from all wikiHow procedures as vector representations. Then, considering the goal inference task, given a step, its corresponding goal, and this goal's embedding, we can search for K goals whose embeddings are the closest to the embedding of the given goal, by a distance metric such as L2 or cosine distance. Though there are millions of goals and steps in wikiHow, we can rely on a fast K-nearest-neighbor search algorithm such as FAISS \cite{johnson2019billion} for efficiency. 

At this point, we have found some highly distracting negative examples, but we still have no guarantee that these negative examples are indeed negative. For the previous example of the goal ``organize receipt'', one of our negative examples could be ``organize receipt for a small business'', which can certainly be the goal of ``store your receipts in a binder''. If we include this case in a multiple-choice example, it would adversely affect our evaluation since the model predictions would have false negatives. There are no easy ways to eliminate these bad distractors. One idea is to check for lexical overlap, and exclude negative examples that are almost identical to the positive example. Another idea is to place more weights on the verb or the object of a goal or a step while performing similarity search. None of these methods are perfectly reliable. 

Fortunately, the noise induced by bad negative examples is not deal-breaking, as we can construct millions of examples from wikiHow, and most of them are good. Hence, models can usually still learn a lot despite some occasional noise. For evaluation, we can be a bit more cautious and run a human validation. For this we can resort to crowdsourcing using Amazon Mechanical Turk. All we need to do is to have people answer our multiple choice questions, and only keep those that most crowd workers answer correctly. In fact, having a large but noisy training set and a small but clean evaluation set is typical in NLP tasks. 

We now have a dataset of multiple choice questions for step and goal inference, but we're not done yet. We have to look out for dataset artifacts, which are spurious statistical correlations that spawn during our data creation process. There are many kind of artifacts, but one notorious kind common in multiple-choice questions comes from negative sampling. Work such as \cite{zellers-etal-2019-hellaswag} has shown that language models can sometimes achieve high performance in multiple-choice questions by only looking at the choices but not the question. This is reminiscent of some badly designed exam questions, where all correct answers are the third one or the longest one. Note that large neural networks work with manifold so that they can pick up complex cues that we human do not intend nor notice, and thus achieve deceptively high performance. Our dataset is no exception. If we mask out the given goal or step but only provide the choices, our best model can achieve significantly over chance accuracy. To deal with this common issue of cues among choices, for each example, we further randomly re-assign a choice as the correct answer. For instance, consider the example with the goal of ``stay healthy'' and 4 candidate steps ``work out'', ``work long hours'', ``try out local restaurants'', and ``eat a lot of food''. The correct answer is apparently ``work out''. To avoid models picking up cues, we then randomly decide that ``try out local restaurants'' is the correct step, while switching the given goal to its corresponding one, ``visit a new city''. This way, if we again mask out the given goal or step but only provide the choices, all models are destined to perform no better than randomly guessing, since all choices have an equal chance of being correct. 

What about the performance? For humans, my collaborator and I each did 100 questions for goal inference and step inference, and we have around 97\% accuracy. In contrast, state-of-the-art language models like RoBERTa \cite{liu2019roberta} can achieve about 85\%. Since it's a 4-choose-1 format, the chance performance is 25\%. This shows that models can quite reliably learn procedural knowledge of goals and steps from our data, which is a good news. 

\subsubsection{Step-Step Temporal Relation} 

Another interesting relation of goals and steps is the step-step temporal relation, namely the order in which steps happen in a procedure. Namely, given a goal and two of its steps, a model needs to choose the one that should happen earlier. Consider this example.
\begin{displayquote}
For the goal ``wash silverware'', which step should happen first?\\
A. Rinse the silverware. B. Dry the silverware.
\end{displayquote}
The answer is unambiguously A. Instead of multiple-choice, the task of step ordering should be formatted as binary classification, since both of the steps and the goal need to be accessible to models. In this case, negative sampling is not an issue, since we can just flip the two steps to form an example with an opposite label. 

For humans, we again did 100 questions for goal inference and step inference, and we again have around 97\% accuracy. In contrast, state-of-the-art language models can achieve about 80\%, with a chance performance of 25\%. 

\subsection{State Tracking} \label{sec:state-tracking}

While events are at the core of natural language, entities are at the core of events. Entities are just things involved in events and they can move around, change form, or remain constant. In procedures and especially instructions, entities are even more crucial, because here the language is usually concise, and every entity mentioned likely plays some important role. For example, to drink bottled water, we can buy a bottle from the vending machine, unscrew the cap, and take a sip. Every entity here is indispensable. Without the vending machine, we would have to procure the water somewhere else; the bottle contains water, and has a cap that must be removed prior to drinking. Hence, the knowledge of what happens to these entities, referred to as their states, is an important knowledge to learn. Such knowledge has plenty of practical uses. For example, it can be used to answer questions about the world (e.g., where is the cap when I'm drinking the water?), or even provide concrete guidance to robots (e.g., if a water-drinking robot knows that the cap is not on the bottle while drinking and that a bottle containing water should be capped, it can infer that it should re-cap the bottle). There are a handful of existing work on state tracking \cite{long-etal-2016-simpler,DBLP:journals/corr/abs-1711-05313,mysore-etal-2019-materials}. We will go over some latest ones. We will focus on the task definition and data instead of methods here.

\begin{figure}[t]
    \centering
    \includegraphics[scale=0.7]{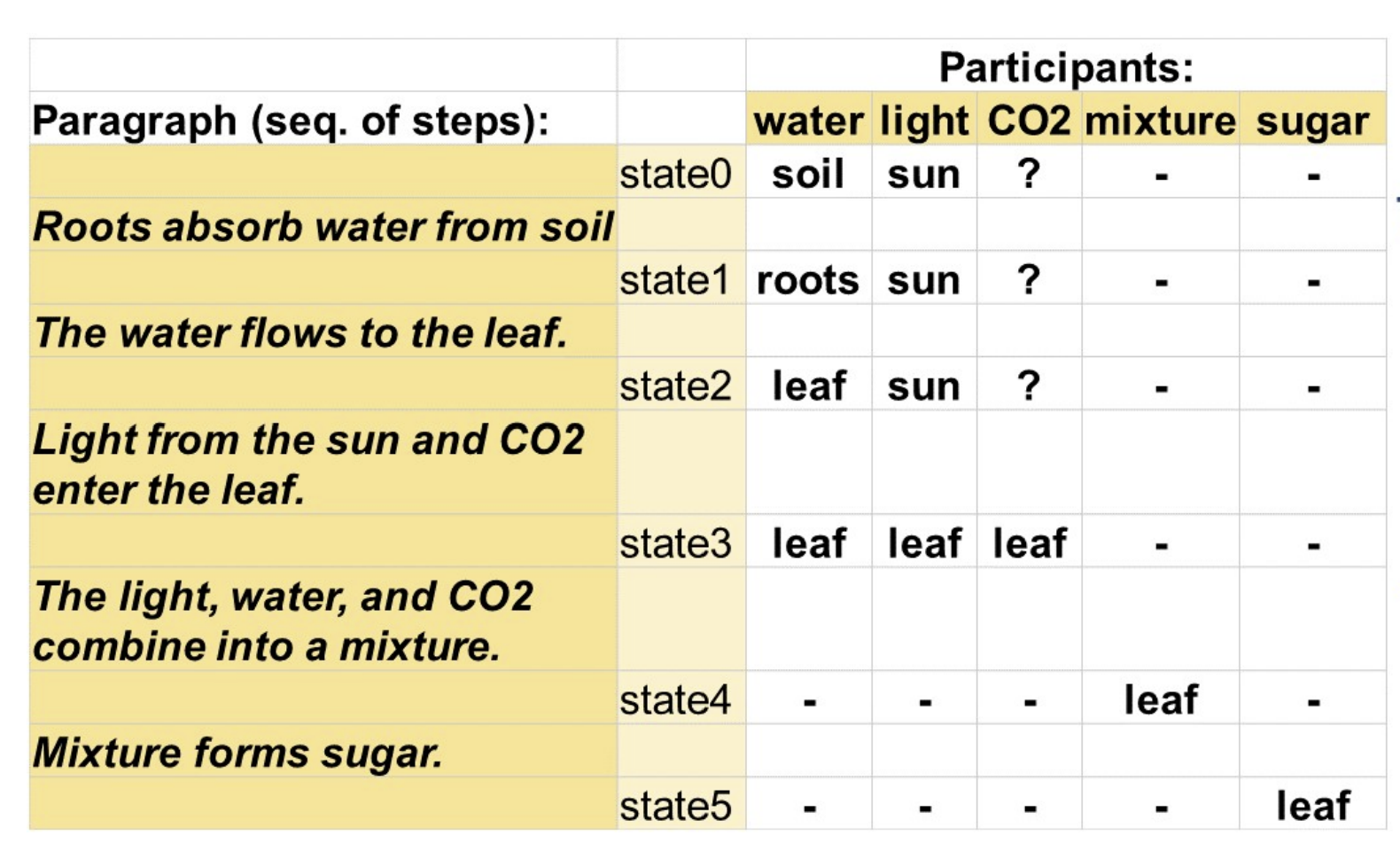}
    \caption{An example of the data annotation in ProPara. Each filled row shows the existence and location of participants between each step (“?” denotes “unknown”, “-” denotes “does not exist”). For example in state0, water is located at the soil.}
    \label{fig:propara}
\end{figure}

\subsubsection{Existence and Location in Scientific Processes}

Let's first look at ProPara \cite{dalvi-etal-2018-tracking}, one of the earliest attempts of state tracking in procedures. The authors of this work focus on a specific kind of procedures: scientific processes (e.g., photosynthesis), unlike the rest of this tutorial which focuses on instructions. They track entity state changes for two attributes: existence and location. Namely, after each step, they want to know if any entity of interest is created or destroyed, and where it is if it exists. See Figure~\ref{fig:propara} for an example. 

To create this dataset, they first come up with a list of prompts of target scientific processes, and then ask crowd workers to write down the steps of these procedures. Next, another group of annotators are asked to write down the entities that have state changes and mark the steps after which they are created or destroyed. Finally, a different group of annotators write down the location of these entities after each step. Thus, a dataset of 81,345 annotations over 488 paragraphs about 183 processes is created. 

The ProPara dataset is great for tracking states. However, it only study two attributes. There are many attributes other than existence and location that have state changes to be tracked (e.g., temperature, openness, color, etc.). Moreover, its domain is limited to scientific processes.

\subsubsection{Open State Tracking in Instructions}
The next work we're going to look at, OpenPI \cite{tandon-etal-2020-dataset}, addresses the previous issues by expanding the attribute set to an open vocabulary. Namely, in addition to existence and location, other attributes like temperature, openness, and color are all fair game. Also, OpenPI gets its data from wikiHow, falling into the line of work on instructions which cover a wide variety of domains. 

\begin{figure}[t]
    \centering
    \includegraphics[scale=0.7]{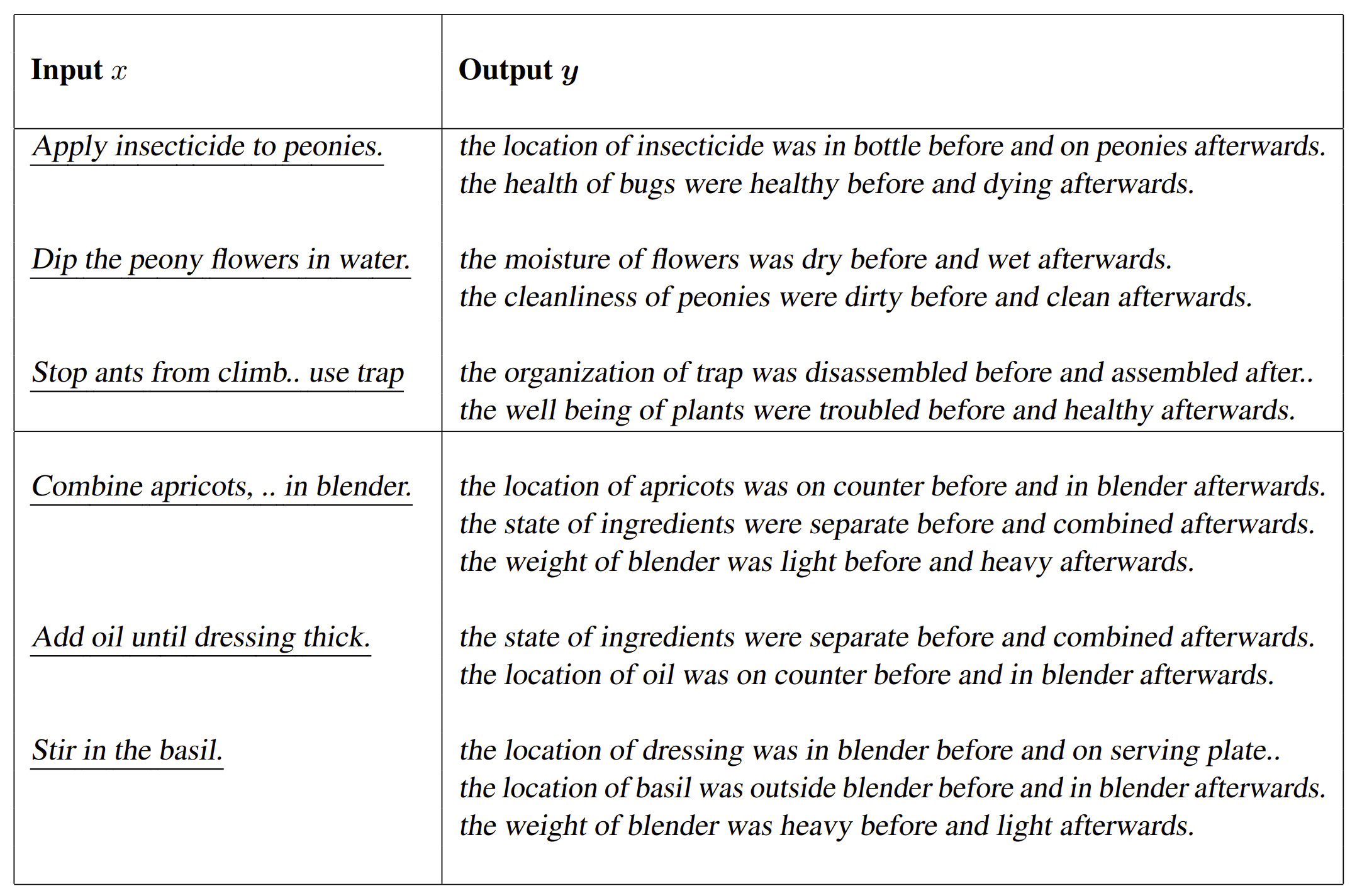}
    \caption{An example of the data annotation in OpenPI. The input comprises a query and a context (past sentences before this step in the paragraph). The output is a set of pre- and post-conditions.}
    \label{fig:openpi}
\end{figure}

Similar to Propara, OpenPI also uses crowdsourcing to collect data. The authors first select a suitable subset of wikiHow articles, and then ask crowd workers to write down a couple of state changes for each step. See an example in Figure~\ref{fig:openpi}. A model trained on this dataset can then also predict some state changes given a step and its contexts. 

This is great, because now a model knows about the entities in a procedure beyond their existence and location. However, there is no guarantee that the state changes written down by the annotators or predicted by the models are complete. For example, during the procedure ``fry fish'', once you ``add the fish to the boiling oil'', many things will happen and many states will change: the fish will become hard and brown, a crust will form on the surface, the oil temperature will temporarily decrease, there will be a sizzling sound, and so much more. Limited by the training data, a model trained on OpenPI is unlikely to predict \textit{all} the state changes. This will cause the model to fail when a query is made for a state that the model does not predict. 

\subsubsection{Question Answering of State Tracking}

The current project of myself and my colleagues takes state tracking in procedures to the next level: what if a model needs to know \textit{every} state change possible in a procedure? Apparently, it is impossible to annotate all state changes. However, we might be able to leverage the state-of-the-art language models' strong few-shot learning ability to figure out these state changes on the fly. For instance, given the first example in Figure~\ref{fig:openpi}, after the step ``apply insecticide to peonies'' a natural question might be ``are the peonies harmful to human body?''

We manually create such a dataset consisting of questions and answers regarding entity state changes, and our preliminary experiments show that a language model trained on OpenPI is unable to answer almost all questions, due to its limit on the predicted state changes. Similarly, strong language models such as T5 also fails almost completely. In contrast, few-shot learners such as GPT3 is able to answer questions regarding entity state changes reasonably. We are currently investigating ways to improve the model, as well as other formulation of questions. 


\section{Applications of Procedures}

Procedures are so ubiquitous in life that knowledge about them can be applied in many scenarios. For example, the technique of reasoning about goals can be used in advertisement to understand people's wants and needs. The technique of reasoning about steps can be used to help users figure out how to do tasks. Moreover, holistically understanding procedures has even more implications. Stories and scripts can be broken down into procedures to be better understood. Schema of procedures can enable comparison and clustering of procedures. 

In this section, we will explore some highly practical applications of procedures.  

\subsection{Suggesting Steps} \label{suggesting-steps}
As humans, we are familiar with some but not all procedures. For example, as a PhD student myself, I know ``how to carry out a research project'' all too well, whereas I struggle at ``how to fly an airplane,'' unlike some of my friends who have had aviation training. It would be nice if someone can show me the steps of tasks that I don't know how to do. Without such someone, my best bet is to look up the tasks online. One major issue of doing so is that there might not be instructions of my task. Moreover, I might require some personalization of the task (e.g., ``fly an airplane while being disabled''). In such cases, it would be helpful if a model can generate steps given a goal, even if it has never seen the goal or the task does not previously exist. Let's see how we can train such a model following \cite{lyu-etal-2021-goal}.

Let's first formally define our task. Given a goal (e.g., ``write a song''), the model should output a sequence of reasonable steps (e.g., ``find inspiration'', ``determine a feel'', ``choose a chord progression'', etc.) that may or may not be ordered. As we will soon see, this task is no easy feat. To do it well, we need holistic understanding of procedures. 

We have some data at our disposal. From wikiHow, we have more than 100k procedures that allow us to perform supervised learning. After splitting them into training and testing sets Let's consider two intuitive approaches for modeling. \\

\minisection{Language Model Generation.} Now is the age for using large language model's generation ability to tackle a variety of tasks. Our task is no exception. Very simply, we can format each procedure as a sequence of the goal and the steps, such as:
\begin{displayquote}
    Sear a steak [SEP] Heat the pan [SEP] Pat dry the steak [SEP] ...
\end{displayquote}
Using a generation model such as T5 or GPT-3, we can then use all the procedures in the training set to finetune the model. During inference, given a goal, the finetuned model would be able to output the steps in a similar format. 

Here is an example of some generated steps for ``icing an injured ankle'' using T5:
\begin{displayquote}
    1. Hang the ankle up in a large, open space.\\
    2. Wrap the injured ankle up in a heated towel.\\
    3. Put the ice pack against your leg for 15-20 minutes.\\
    4. Elevate your legs above the level of irritation.\\
    5. Take an over-the-counter pain reliever.\\
    6. See a doctor if 2-3 days persist or worsen.\\
    7. Keep your foot elevated to avoid further swelling.\\
    8. Use an orthopedic shoes to trim your ankle.\\
    9. Wear the right shoes and footware.
\end{displayquote}

Are these suggested steps any good? At a quick glance, they may look coherent, but at a closer look, almost all steps are problematic and are no way to really treat your injured ankle. Here are some issues that are in fact representative of language model generation's common failures. 

\begin{enumerate}
    \item Illogical actions. In 1, you don't need a large, open space to hang your ankle. In 8, you probably don't want to make things worse by ``trimming'' your ankle. 
    \item Irrelevant steps. In 2, heated towel has nothing to do with icing. In 4, it would be strange to refer to an injury as ``irritation''. 
    \item Grammatical issues. In 6 and 8, the grammar is incorrect. 
\end{enumerate}

While end-to-end language models are very straightforward to use, the ways that they can be improved are limited. Naturally, we may adjust the language model's hyperparameters or configurations such as temperature. We may also try different data format, or experiment with data distillation, augmentation, or curriculum learning. However, I myself don't see these tweaks bring about large improvements. Most likely, some insights into the task of suggesting steps is need to handle it better. \\

\minisection{Inferring and then Ordering Steps.} Instead of the end-to-end approach above, we could also reuse our model that infers steps given a goal, and the one that orders steps given a goal. Recall that the step inference model is a transformer (such as RoBERTa) that takes in two events and calculates a probability of a goal-step relation. To suggest steps given a goal, we can then use this model to calculate the pairwise probability between the goal and every single possible event, and keep the events with the highest probability of actually being the step of the given goal. While we can't possibly get all events, we can use all the steps from all articles in wikiHow as a candidate pool. Using this approach, we have already eliminated issue 1 and 3 of language model generation, given that wikiHow steps are grammatical and sensical. 

Suppose we now have the pairwise goal-step probability between the given goal and each wikiHow step. We can then choose top K steps with the highest probability, where K is a design choice and depends on how long we want the suggestion to be. 

Next, we can order these K steps by running our step ordering model, a transformer that takes in two steps and a goal and calculates the probability of the first step preceding the second. To use this model to order or rank more than 2 steps, we can first order every pair of steps, and then poll the number of times a step precedes any other step. This is reminiscent of a sports tournament where every two players or teams play, and the ranking is determined by the number of wins each player or team ends up with. 

Before looking at some examples of model generation, let's think about the pros and cons of this approach. On the bright side, as mentioned before, since all steps come from wikiHow which has an editorial process, most of them will be logical and well-formed. However, there are also many possible failures. First, the step inference model is not perfect, and some non-steps, perhaps irrelevant ones, might be mistakenly ranked high. Even if the step inference model were perfect, the top-ranked steps are likely to contain duplicates. Consider the goal ``choose a laptop'', and a reasonable step ``decide on a budget.'' It is obvious that there could be many steps in wikiHow about deciding on budgets from articles of buying different things. These steps might also have different wordings, such as ``choose a reasonable budget'', ``set a budget'', ``decide on how much you want to spend'', etc. A perfect step inference model is bound to score all of these steps high, so that after keeping the top K, we will end up with many steps with the same meaning. 

To ameliorate this issue, we can leverage a paraphrase detection module which was not used in \cite{lyu-etal-2021-goal}. There are many options. Simply, we can use a sentence embedding model such as sBERT finetuned on some paraphrase corpus to calculate the embedding for each candidate step. Then, we can use algorithms such as K-means to cluster the steps by their meaning, before we choose one step from each cluster. Alternatively, we can also use a dedicated paraphrase detection model such as \cite{nighojkar-licato-2021-improving} to calculate pairwise similarity among the candidate steps. 

We have looked at possible failures during the phrase of inferring steps, and will now look at those during step ordering. Just like step inference, the step ordering model is imperfect and would make erroneous predictions. Moreover, such errors can propagate via the tournament ranking algorithm. For example, given 3 candidate steps A, B, and C, if the step ordering model predicts that A precedes B, B precedes C, and C precedes A, an uninformed tie-breaking mechanism (such as randomly choosing one) needs to be invoked. 

One glaring weakness of our current temporal ordering approach is that we rely on aggregating pairwise predictions. A more reasonable alternative may be to directly train a generation model to re-order the entire sequence of steps. To the best of my knowledge, this is only attempted by one piece of work \cite{byrne-etal-2021-tickettalk}, which learns a denoising autoencoder with event sequences that have been shuffled. A minor difference between their work and our task is that their model is designed for narrative events which do not necessarily have a goal, like procedural events. 

Let's look at an example set of steps predicted by our approach, given the same goal as above ``ice an injured ankle''. 

\begin{displayquote}
1. Place an ice pack on the injured ankle for 15-20 minutes.\\
2. Rest the injured ankle as much as possible for 48 hours.\\
3. Compress the injured ankle for 48 hours to prevent swelling.\\
4. Elevate the injured ankle above your heart to decrease any swelling.\\
5. Go to the emergency room if the injured ankle cannot bear weight.\\
6. Apply a compression bandage to a sprained ankle.\\
7. Consult with a doctor if the ankle does not improve after 2-3 days.\\
8. Ask a friend to grab your ankles from the bottom.
\end{displayquote}
Looks pretty good! Here are some common failures.
\begin{enumerate}
    \item Irrelevant steps. Step 8 may be irrelevant, but we don't have enough information.
    \item Duplicated steps. In this case, we do not observe much duplication within the steps.
    \item Wrong ordering. Step 6 should happen as early as step 3, while step 5 should be the very first thing to do.
\end{enumerate}

Until now, we have looked at some models to suggest steps given a goal and some of their sample output but we haven't discussed how we can systematically evaluate the suggested steps. Similar to many language generation tasks, faithful automatic evaluation is hard to come by. The key issue is that there is no single right answer, and no straightforward way to compute the similarity between steps suggested by a model and the reference steps. According to \cite{lyu-etal-2021-goal}, common metrics such as BERTScore \cite{zhang2019bertscore} have almost no correlation with human judgements. 

For now, we're better off relying on human judgement, designing which is non-trivial either. For procedures that most of us know about, we can perform human judgement without a reference set of steps. One systematic way to score a sequence of suggested step is to \textit{edit} the steps, by either adding, removing, or re-ordering steps. The number of such edit operations is called \textit{edit distance}. Intuitively, well-suggested steps need few edits, thus having a small edit distance, and vice versa. Note that such evaluation process has its share of shortcomings. For example, it is quite subjective to add steps, as people may expect different level of details or granularity from procedures. 

As a final remark, suggesting steps given a goal is an important application of procedural knowledge. While current models can do this to some extent, there is still a large room of improvement. 

\subsection{Finding Sub-Steps} \label{sec:substeps}
A procedure consists of a goal and some steps. In fact, each step can itself represent another procedure. For example, the procedure ``make a movie'' may have a step ``buy a camera'', which in turn can be the goal of another procedure consisting of steps such as ``set a budget'', ``decide use case'', ``read reviews'', etc. By this logic, procedures are \textbf{hierarchical}, and steps may have sub-steps. This is also the case for general, narrative events, and such hierarchical relations have been studied \cite{bisk-etal-2019-benchmarking} in some but not much work. 

In the previous section we touched on the issue of event granularity in procedures. When it comes to procedures, the level of details in the steps vary greatly. To teach an adept iPhone user how to ``fix a frozen iPhone X'', you may just say ``force restart it'', resulting in a 1-step procedure. For general audience, you may say ``press the volume up botton'', ``press the volume down botton'', and ``hold the volume up botton'', resulting in a 3-step procedure. To teach someone who has never used electronics, you might need to further explain how to hold buttons and where they are. Having control over the granularity of procedures is important in many ways. Not only does it give us control over the level of details in instructions, but it also influences some downstream tasks as we will see later. Such granularity can be easily accessed if we have a hierarchy of procedures, where abstractions are higher and details are lower in the hierarchy. 

\begin{figure}[t]
    \centering
    \includegraphics[width=0.9\textwidth]{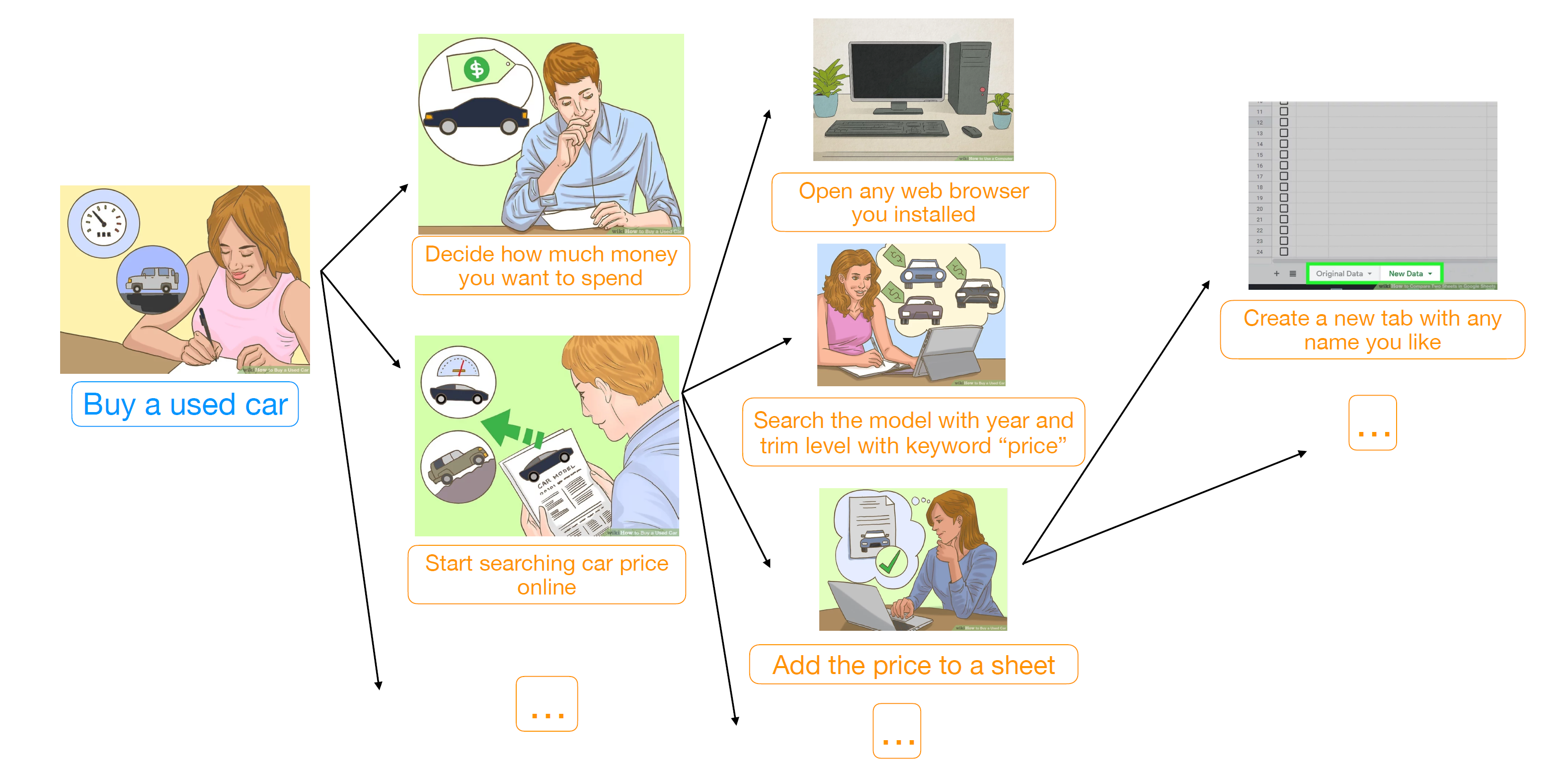}
    \caption{Example of procedural hierarchies in wikiHow.}
    \label{fig:wikihow_hierarchy}
\end{figure}

Building a procedural hierarchy is equivalent to recursively finding sub-steps for steps, which is in turn equivalent to \textbf{linking steps to procedures}. In wikiHow, there exists some hyperlinks from a step to another article. For example, in ``How to Make a YouTube Video'', the step ``Record content that is on your computer monitor'' is linked to another wikiHow article ``How to Record Your Computer Screen''. Unfortunately, such hyperlinks are very scarce. If all steps were to have such links when applicable, we can then find sub-steps trivially, and thus a procedural hierarchy can be constructed. In fact, linking contents in wikiHow has been attempted in previous years \cite{10.1145/2567948.2578846,10.1145/2872518.2890585}. Let's see how we can use state-of-the-art NLP techniques to tackle this problem in one of my own work \cite{zhou-etal-2022-show}.

Our task is to link all steps in wikiHow to some new article or procedure when possible. Sometimes, a step can be too specific (e.g., ``whisk together two eggs and a cup of heavy cream'') or too low-level (e.g., ``breathe'') to have sub-steps, in which case we mark this step as unlinkable. For now, let's focus on linking whole steps in wikiHow, though it's also reasonable to link a part of the step, such as individual actions. 

Our labeled data comes from wikiHow's existing hyperlinks, which are provided by editors and high-quality in general. We split this dataset into training and testing splits. Now, given a step, how do we find a corresponding wikiHow article? This is a typical information retrieval or search problem, for which sentence embeddings and semantic similarity are typical solutions. For now, let's only try matching each step with the goals of the articles, ignoring other information in the articles. We have two modeling options. 
\begin{enumerate}
    \item \textbf{Unsupervised.} We can directly use a pretrained sentence embedding model such as sBERT to convert each wikiHow step and goal into a vector representation. Then, we can use a fast K-nearest-neighbor (KNN) algorithm such as FAISS for find the closest goal of each step.
    \item \textbf{Supervised.} We can train a network to take in a step and a goal, embed them using sBERT, and predict whether they should be linked. In our training data of hyperlinks, each training instance is a step and the goal of the article that the step is linked to. This pair of step and goal should have the same meaning. 
\end{enumerate}

These two methods each have their pros and cons. The unsupervised method is fast thanks to the efficient KNN search, but fails to leverage our training data, leading to lower accuracy. The supervised method is slow because each step-goal pair needs to be computed, but has better performance by learning from step-goal pairs that have the same meaning. Fortunately we don't have to choose. It is conventional in information retrieval tasks to sequentially perform these two methods in a \textbf{retrieve-then-rerank} pipeline. First, we use the unsupervised method to narrow down the candidate goals. Then, we use the supervised method to re-score each pair of the step and a candidate goal, and re-rank the goals. 

You can see a demo of the resulting procedural hierarchy at \url{https://wikihow-hierarchy.github.io/}.

\subsection{Guiding Dialogues} 

Procedural knowledge has seen use in commercial voice assistants such as Amazon Alexa. At the time of writing, if you ask Alexa how to do something, it will guide you through a corresponding wikiHow article or a Whole Foods Market recipe\footnote{\url{https://www.wholefoodsmarket.com/recipes}} step by step. Navigating through the steps itself is straightforward, but almost any other interesting interaction require procedural knowledge. For example, if a user asks for clarification of a step, then the system must know what article to redirect the user to given the contexts (Section~\ref{sec:substeps}). If a user asks a question during the procedure, knowledge about entity states might be needed to provide an answer (Section~\ref{sec:state-tracking}). If a user run into issues such as not having a certain tool or ingredient, the system needs to invoke what-if reasoning. More futuristically, if a user's desired task does not have a good set of instructions available, the system can suggest some reasonable steps (Section~\ref{suggesting-steps}). In fact, our team at the University of Pennsylvania participated in the Alexa Prize TaskBot Challenge\footnote{\url{https://www.amazon.science/alexa-prize}} from 2021 to 2022, and worked on most of the above aspects. 

What we have touched upon are just a tip of the iceberg of what procedural knowledge can be used to do in real life. I can't prove this claim to you at this time, because most of these potential applications remain ongoing or unexplored work. For example, procedural knowledge can be used to generate texts, such as a story about going through a lawsuit, or a fictional narrative about scaling a castle wall. I'll happily leave all those to future work.





\section{Procedure and Script}
\cite{schank_scripts_1977} introduced the concept of \textbf{script} as a way to unify knowledge of events, reconciling distinctive conventions from psychology, cognitive science, and of course, artificial intelligence. From the perspective of AI, \textbf{script learning} is an ambitious journey towards artificial general intelligence, as it attempts to reduce arbitrarily complex human activities into low-level, symbolic representations that machines can conceivably understand. Such an attempt was tremendously difficult, as there are myriad of exceptions and edge cases that any intelligent system must account for. Thus, script learning culminated in popularity around that time, but was thwarted by the limitation of technology. In recent years with the advances in technology, it has been steadily picking up traction again. Procedures and scripts share a lot of common grounds, since they are both semantic constructs that describe events. In this section, we will juxtapose the two concepts and gain some interesting insights.

According to Schank, 
\begin{displayquote}
A script is a structure that describes appropriate sequences of events in a particular context.
\end{displayquote}
For example, the following is a script.
\begin{displayquote}
John went to a restaurant. He asked the waitress for coq au vin. He paid the check and left.
\end{displayquote}
The following is not a script.
\begin{displayquote}
John was walking on the street. He thought of cabbages. He
picked up a shoe horn.
\end{displayquote}

An astute reader can immediately notice how a script is similar to a procedure. The only difference is the emphasis on the ``appropriateness'' and the ``context'', and the lack of the emphasis on the ``goal'' as in a procedure. Indeed, a procedure can be seen as a special kind of script: a much simpler one at that. In fact, Schank defined a special type of script, called \textbf{instrumental script}:
\begin{displayquote}
The crucial differences between instrumental and situational scripts are with respect to the number of actors, and the overall \textbf{intention} or \textbf{goal} of the script. 
\end{displayquote}

This looks just like procedures\footnote{There are still some differences. For example, Schank claims that instrumental scripts have ``very rigid ordering'', and ``each action must be done.'' While these are sometimes true, we have seen that steps in a procedure often have flexible local ordering, and varying degree of importance.}. Conversely, other types of scripts, such as situational script containing sequences of actions in certain scenarios, are much more full-fledged, with many properties that are absent in procedures. Note again that the term ``procedure'' discussed here mainly points to what has been studied in NLP (e.g., natural language instructions).

Let's compare some features of a script and a procedure. 

\minisection {A script can have multiple explicit actors. A procedure usually has only one implicit actor.} 
A script can involve more than one person, interacting with each other in an arbitrarily complex fashion. Each person has different viewpoints, and thus different scripts from different perspectives. However, most procedures, especially those in the form of instructions (e.g., recipes, how-to guide) do not have any explicit actor. The actor of the steps is implied to be the reader or the person following these instructions. Nonetheless, it is important to recall that instructions are merely a primary focus for research on procedures, and procedures do not necessarily appear in the form of instructions. Conceivable, there could be a procedure where two person collaborates to achieve some goal. In such case, inter-actor interactions present in scripts can also be present in procedures.

\minisection {A script can meaningfully contain multiple procedures.} 
First, it is crucial to note that both scripts and procedures can be hierarchical. Namely, a script or procedure (e.g., eat at a restaurant) can contain other scripts or procedures (e.g., order food, paying the check). Since a procedure (as defined in this tutorial) is centered around one goal, it is logical to assert that its sub-procedures are thus centered around its sub-goals. However, a script can involve no goal, one goal, or multiple goals. These goals can belong to different actors, one actor but at different time stage, or one actor simultaneously. Naturally, the events in the script that serve any particular goal would form a procedure. Thus, a script can meaningfully contain multiple procedures. The reverse can also be true, though usually less meaningful. For example, the procedure of ``getting a college degree'' might have a step ``take classes'', which could be represented by a script with all the complexities such as multiple actors, sub-scripts, etc. However, since one would only care about events in this script that serves the goal ``take classes'' or its sub-goals, it is much convenient to view the step ``take classes'' as another procedure rather than a full-fledged script.

\minisection {A script is subject to interferences and distractions, while a procedure usually is not.} One of the most important distinction between scripts and procedures is that scripts often interact and intertwine with each other, while procedures often do not, barring hierarchical relations. For example, during a script of ``working at a computer at home'', we inattentive mortals can easily be distracted by another script ``watching some YouTube videos'', or be interfered by another script ``attend to the ringing doorbell''. When such sidetracking happens, script learning becomes extremely complex. What will the actor do? Do they switch to the other script and just forfeit the current script? Do they deal with the other script and return? Are there influences or damages to the current script caused by the sidetracking? How will the actor deal with that? All these are beyond our scope, since working with procedures we opt to mostly not deal with them. For example, instructions such as recipes rarely talk about what happens if things go wrong, or if you suddenly decide to cook another dish. They are locked in onto the task at hand. However, interferences and distractions definitely \textit{can} happen in procedures. For example, what would happen if you don't have certain ingredients? Your produce goes bad? Your house catch fire (god forbid)? This research question remains largely unexplored.

The key takeaway here is that procedures are a specific type of scripts,  with significantly lower complexity and variation. Scripts are challenging to learn; procedures are relatively simpler.

\section{Recipes}
Just like procedures are simpler scripts, recipes are simpler procedures. A strand of work on procedures focuses exclusively on recipes \cite{tasse2008sour,10.1145/355214.355237,mori-etal-2014-flow,kiddon-etal-2015-mise,7574705,zhou2018towards,bisk-etal-2019-benchmarking,lin-etal-2020-recipe,donatelli-etal-2021-aligning}, for good reasons. A special subset of instructions, recipes are abundant and well-structured. They also uses regular, clear, and concise language. They have a small scope of topics (i.e., food) and involved actions and entities. In fact, much previous work on recipes touch on the same aspects as we have seen in this tutorial regarding general procedures. An interesting question is to what extent the data, methods, and findings of these pieces of work apply to general procedures. In this section, we will examine several more influential pieces of work. 

\cite{mori-etal-2014-flow} tries to represent recipes with a flow graph, comparable to what we have seen for general procedures in Section~\ref{sec:representation}. The authors of this work obtain recipes from a recipe site, while one can of course obtain other instructions from sites such as wikiHow. They represent recipes as a graph, where vertices are food, tools, or actions. Naturally, this can be generalized to any entities and events in general procedures. The edges stand for relations between entities, such as equality, subset, application, etc. None of those are really specific to recipes, and can likely apply to any procedure. 

\cite{bisk-etal-2019-benchmarking} collects a dataset with hierarchical cooking actions with different granularity. Concretely, the authors ask crowd worker to explain captions from cooking videos in simple terms, resulting in lower-level sub-steps. The same methodology can be used for general procedures using captions from datasets such as HowTo100M \cite{miech2019howto100m}. Then, they propose a cloze task by having models predict some omitted actions. This naturally applies to general procedures too. 

\cite{lin-etal-2020-recipe} and \cite{donatelli-etal-2021-aligning} align related or equivalent actions between two dish recipes of the same dish. Similar to the cases above, the authors' choice of working on recipes instead of general procedures is motivated the existing datasets. Their method of alignment also only assumes linguistic features of instructional language, and not just recipes. 

Without scrutinizing every single work on recipes, it does seem that most if not all of these pieces of work can be easily generalized to general procedures, especially instructions of non-cooking tasks. However, whether this is the case still requires investigation. Recipes may oversimplify things. For example, in a recipe there is only one person involved (i.e., the chef), while in other procedures there might be several (e.g., call the hotel representative to ask for availability). Having multiple agents in a procedure might require knowledge about interaction in different scenarios. Also, most entities in recipes are either ingredients or tools, while in general procedures they might vary a lot more (e.g., locales, institutions, etc.). The flow and expressions in recipes are usually more straightforward, while in other procedures there might be more complexity (e.g., conditionals, explanations, background information, etc.). In summary, I believe that future work should attempt consider general procedures including but not limited to recipes, given the abundant existing web resources. 

\section{Recap}
In this tutorial, we have gone through research on reasoning about procedures in the NLP community. We started by appreciating this trendy line of research and briefly learning its history. Next, we saw that procedural data used to come from human annotation, but now can also be obtained to-scale from web resources. We then looked at some efforts to represent procedures. At the heart of each of these representations is granular knowledge regarding components of procedures. We also studied some work on learning such knowledge. Then, we examined some practical applications of procedures, with said knowledge equipped. Finally, we compare procedures with scripts, a superset of procedures, and recipes, a subset of procedures, both of which have also been receiving great attention from the NLP community. 

\bibliographystyle{apalike}
{\footnotesize
\bibliography{anthology,custom}}

\begin{thebibliography}{}

\bibitem[Addis and Borrajo, 2011]{Addis2011FromUW}
Addis, A. and Borrajo, D. (2011).
\newblock From unstructured web knowledge to plan descriptions.
\newblock In {\em Information Retrieval and Mining in Distributed
  Environments}.

\bibitem[Ahn et~al., 2022]{ahn2022can}
Ahn, M., Brohan, A., Brown, N., Chebotar, Y., Cortes, O., David, B., Finn, C.,
  Gopalakrishnan, K., Hausman, K., Herzog, A., et~al. (2022).
\newblock Do as i can, not as i say: Grounding language in robotic affordances.
\newblock {\em arXiv preprint arXiv:2204.01691}.

\bibitem[Aouladomar, 2005]{aouladomar2005preliminary}
Aouladomar, F. (2005).
\newblock A preliminary analysis of the discursive and rhetorical structure of
  procedural texts.
\newblock In {\em Symposium on the Exploration and Modeling of Meaning}.
  Citeseer.

\bibitem[Artzi and Zettlemoyer, 2013]{artzi-zettlemoyer-2013-weakly}
Artzi, Y. and Zettlemoyer, L. (2013).
\newblock Weakly supervised learning of semantic parsers for mapping
  instructions to actions.
\newblock {\em Transactions of the Association for Computational Linguistics},
  1:49--62.

\bibitem[Bielsa and Donnell, 2002]{bielsa2002semantic}
Bielsa, S. and Donnell, M. (2002).
\newblock Semantic functions in instructional texts: A comparison betw een
  english and spanish.
\newblock In {\em Proceedings of the 2nd International Contrastive Linguistics
  Conference}, pages 723--732.

\bibitem[Bisk et~al., 2019]{bisk-etal-2019-benchmarking}
Bisk, Y., Buys, J., Pichotta, K., and Choi, Y. (2019).
\newblock Benchmarking hierarchical script knowledge.
\newblock In {\em Proceedings of the 2019 Conference of the North {A}merican
  Chapter of the Association for Computational Linguistics: Human Language
  Technologies, Volume 1 (Long and Short Papers)}, pages 4077--4085,
  Minneapolis, Minnesota. Association for Computational Linguistics.

\bibitem[Bosselut et~al., 2017]{DBLP:journals/corr/abs-1711-05313}
Bosselut, A., Levy, O., Holtzman, A., Ennis, C., Fox, D., and Choi, Y. (2017).
\newblock Simulating action dynamics with neural process networks.
\newblock {\em CoRR}, abs/1711.05313.

\bibitem[Branavan et~al., 2009]{branavan-etal-2009-reinforcement}
Branavan, S., Chen, H., Zettlemoyer, L., and Barzilay, R. (2009).
\newblock Reinforcement learning for mapping instructions to actions.
\newblock In {\em Proceedings of the Joint Conference of the 47th Annual
  Meeting of the {ACL} and the 4th International Joint Conference on Natural
  Language Processing of the {AFNLP}}, pages 82--90, Suntec, Singapore.
  Association for Computational Linguistics.

\bibitem[Brown et~al., 2020a]{NEURIPS2020_1457c0d6}
Brown, T., Mann, B., Ryder, N., Subbiah, M., Kaplan, J.~D., Dhariwal, P.,
  Neelakantan, A., Shyam, P., Sastry, G., Askell, A., Agarwal, S.,
  Herbert-Voss, A., Krueger, G., Henighan, T., Child, R., Ramesh, A., Ziegler,
  D., Wu, J., Winter, C., Hesse, C., Chen, M., Sigler, E., Litwin, M., Gray,
  S., Chess, B., Clark, J., Berner, C., McCandlish, S., Radford, A., Sutskever,
  I., and Amodei, D. (2020a).
\newblock Language models are few-shot learners.
\newblock In Larochelle, H., Ranzato, M., Hadsell, R., Balcan, M.~F., and Lin,
  H., editors, {\em Advances in Neural Information Processing Systems},
  volume~33, pages 1877--1901. Curran Associates, Inc.

\bibitem[Brown et~al., 2020b]{brown2020language}
Brown, T., Mann, B., Ryder, N., Subbiah, M., Kaplan, J.~D., Dhariwal, P.,
  Neelakantan, A., Shyam, P., Sastry, G., Askell, A., et~al. (2020b).
\newblock Language models are few-shot learners.
\newblock {\em Advances in neural information processing systems},
  33:1877--1901.

\bibitem[Byrne et~al., 2021]{byrne-etal-2021-tickettalk}
Byrne, B., Krishnamoorthi, K., Ganesh, S., and Kale, M. (2021).
\newblock {T}icket{T}alk: Toward human-level performance with end-to-end,
  transaction-based dialog systems.
\newblock In {\em Proceedings of the 59th Annual Meeting of the Association for
  Computational Linguistics and the 11th International Joint Conference on
  Natural Language Processing (Volume 1: Long Papers)}, pages 671--680, Online.
  Association for Computational Linguistics.

\bibitem[Chen and Mooney, 2011]{10.5555/2900423.2900560}
Chen, D.~L. and Mooney, R.~J. (2011).
\newblock Learning to interpret natural language navigation instructions from
  observations.
\newblock In {\em Proceedings of the Twenty-Fifth AAAI Conference on Artificial
  Intelligence}, AAAI'11, page 859–865. AAAI Press.

\bibitem[Chen et~al., 2021]{chen-etal-2021-event}
Chen, M., Zhang, H., Ning, Q., Li, M., Ji, H., McKeown, K., and Roth, D.
  (2021).
\newblock Event-centric natural language processing.
\newblock In {\em Proceedings of the 59th Annual Meeting of the Association for
  Computational Linguistics and the 11th International Joint Conference on
  Natural Language Processing: Tutorial Abstracts}, pages 6--14, Online.
  Association for Computational Linguistics.

\bibitem[Chen, 1976]{10.1145/320434.320440}
Chen, P. P.-S. (1976).
\newblock The entity-relationship model—toward a unified view of data.
\newblock {\em ACM Trans. Database Syst.}, 1(1):9–36.

\bibitem[Chernov et~al., 2016]{10.1145/2872518.2890585}
Chernov, A., Lagos, N., Gall\'{e}, M., and S\'{a}ndor, A. (2016).
\newblock Enriching how-to guides by linking actionable phrases.
\newblock In {\em Proceedings of the 25th International Conference Companion on
  World Wide Web}, WWW '16 Companion, page 939–944, Republic and Canton of
  Geneva, CHE. International World Wide Web Conferences Steering Committee.

\bibitem[Dalvi et~al., 2018]{dalvi-etal-2018-tracking}
Dalvi, B., Huang, L., Tandon, N., Yih, W.-t., and Clark, P. (2018).
\newblock Tracking state changes in procedural text: a challenge dataset and
  models for process paragraph comprehension.
\newblock In {\em Proceedings of the 2018 Conference of the North {A}merican
  Chapter of the Association for Computational Linguistics: Human Language
  Technologies, Volume 1 (Long Papers)}, pages 1595--1604, New Orleans,
  Louisiana. Association for Computational Linguistics.

\bibitem[Dalvi et~al., 2019]{dalvi-etal-2019-everything}
Dalvi, B., Tandon, N., Bosselut, A., Yih, W.-t., and Clark, P. (2019).
\newblock Everything happens for a reason: Discovering the purpose of actions
  in procedural text.
\newblock In {\em Proceedings of the 2019 Conference on Empirical Methods in
  Natural Language Processing and the 9th International Joint Conference on
  Natural Language Processing (EMNLP-IJCNLP)}, pages 4496--4505, Hong Kong,
  China. Association for Computational Linguistics.

\bibitem[de~Rijke et~al., 2005]{deRijke2005QuestionAW}
de~Rijke, M., Bunt, H., Geertzen, J., and Thijsse, E. (2005).
\newblock Question answering: What's next?

\bibitem[Delpech and Saint-Dizier,
  2008]{delpech-saint-dizier-2008-investigating}
Delpech, E. and Saint-Dizier, P. (2008).
\newblock Investigating the structure of procedural texts for answering how-to
  questions.
\newblock In {\em Proceedings of the Sixth International Conference on Language
  Resources and Evaluation ({LREC}'08)}, Marrakech, Morocco. European Language
  Resources Association (ELRA).

\bibitem[Devlin et~al., 2019]{devlin-etal-2019-bert}
Devlin, J., Chang, M.-W., Lee, K., and Toutanova, K. (2019).
\newblock {BERT}: Pre-training of deep bidirectional transformers for language
  understanding.
\newblock In {\em Proceedings of the 2019 Conference of the North {A}merican
  Chapter of the Association for Computational Linguistics: Human Language
  Technologies, Volume 1 (Long and Short Papers)}, pages 4171--4186,
  Minneapolis, Minnesota. Association for Computational Linguistics.

\bibitem[Doddington et~al., 2004]{doddington-etal-2004-automatic}
Doddington, G., Mitchell, A., Przybocki, M., Ramshaw, L., Strassel, S., and
  Weischedel, R. (2004).
\newblock The automatic content extraction ({ACE}) program {--} tasks, data,
  and evaluation.
\newblock In {\em Proceedings of the Fourth International Conference on
  Language Resources and Evaluation ({LREC}{'}04)}, Lisbon, Portugal. European
  Language Resources Association (ELRA).

\bibitem[Donatelli et~al., 2021]{donatelli-etal-2021-aligning}
Donatelli, L., Schmidt, T., Biswas, D., K{\"o}hn, A., Zhai, F., and Koller, A.
  (2021).
\newblock Aligning actions across recipe graphs.
\newblock In {\em Proceedings of the 2021 Conference on Empirical Methods in
  Natural Language Processing}, pages 6930--6942, Online and Punta Cana,
  Dominican Republic. Association for Computational Linguistics.

\bibitem[Ellis et~al., 2014]{ellis2014overview}
Ellis, J., Getman, J., and Strassel, S.~M. (2014).
\newblock Overview of linguistic resources for the tac kbp 2014 evaluations:
  Planning, execution, and results.
\newblock In {\em Proceedings of TAC KBP 2014 Workshop, National Institute of
  Standards and Technology}, pages 17--18.

\bibitem[Fritz and Gil, 2011]{fritz2011formal}
Fritz, C. and Gil, Y. (2011).
\newblock A formal framework for combining natural instruction and
  demonstration for end-user programming.
\newblock In {\em Proceedings of the 16th international conference on
  intelligent user interfaces}, pages 237--246.

\bibitem[Gil, 2015]{10.1145/2531920}
Gil, Y. (2015).
\newblock Human tutorial instruction in the raw.
\newblock {\em ACM Trans. Interact. Intell. Syst.}, 5(1).

\bibitem[Gil et~al., 2011]{gil2011tellme}
Gil, Y., Ratnakar, V., and Frtiz, C. (2011).
\newblock Tellme: Learning procedures from tutorial instruction.
\newblock In {\em Proceedings of the 16th international conference on
  intelligent user interfaces}, pages 227--236.

\bibitem[Hamada et~al., 2000]{10.1145/355214.355237}
Hamada, R., Ide, I., Sakai, S., and Tanaka, H. (2000).
\newblock Structural analysis of cooking preparation steps in japanese.
\newblock In {\em Proceedings of the Fifth International Workshop on on
  Information Retrieval with Asian Languages}, IRAL '00, page 157–164, New
  York, NY, USA. Association for Computing Machinery.

\bibitem[Han et~al., 2019]{han-etal-2019-joint}
Han, R., Ning, Q., and Peng, N. (2019).
\newblock Joint event and temporal relation extraction with shared
  representations and structured prediction.
\newblock In {\em Proceedings of the 2019 Conference on Empirical Methods in
  Natural Language Processing and the 9th International Joint Conference on
  Natural Language Processing (EMNLP-IJCNLP)}, pages 434--444, Hong Kong,
  China. Association for Computational Linguistics.

\bibitem[Huang et~al., 2021]{NEURIPS2021_d367eef1}
Huang, J., Li, Z., Chen, B., Samel, K., Naik, M., Song, L., and Si, X. (2021).
\newblock Scallop: From probabilistic deductive databases to scalable
  differentiable reasoning.
\newblock In Ranzato, M., Beygelzimer, A., Dauphin, Y., Liang, P., and Vaughan,
  J.~W., editors, {\em Advances in Neural Information Processing Systems},
  volume~34, pages 25134--25145. Curran Associates, Inc.

\bibitem[Huang et~al., 2022]{huang2022language}
Huang, W., Abbeel, P., Pathak, D., and Mordatch, I. (2022).
\newblock Language models as zero-shot planners: Extracting actionable
  knowledge for embodied agents.
\newblock {\em arXiv preprint arXiv:2201.07207}.

\bibitem[Johnson et~al., 2019]{johnson2019billion}
Johnson, J., Douze, M., and J{\'e}gou, H. (2019).
\newblock Billion-scale similarity search with {GPUs}.
\newblock {\em IEEE Transactions on Big Data}, 7(3):535--547.

\bibitem[Kiddon et~al., 2015]{kiddon-etal-2015-mise}
Kiddon, C., Ponnuraj, G.~T., Zettlemoyer, L., and Choi, Y. (2015).
\newblock Mise en place: Unsupervised interpretation of instructional recipes.
\newblock In {\em Proceedings of the 2015 Conference on Empirical Methods in
  Natural Language Processing}, pages 982--992, Lisbon, Portugal. Association
  for Computational Linguistics.

\bibitem[Kosseim and Lapalme, 1994]{kosseim1994content}
Kosseim, L. and Lapalme, G. (1994).
\newblock Content and rhetorical status selection in instructional texts.
\newblock In {\em Proceedings of the Seventh International Workshop on Natural
  Language Generation}.

\bibitem[Kosseim and Lapalme, 2000]{https://doi.org/10.1111/0824-7935.00118}
Kosseim, L. and Lapalme, G. (2000).
\newblock Choosing rhetorical structures to plan instructional texts.
\newblock {\em Computational Intelligence}, 16(3):408--445.

\bibitem[Koupaee and Wang, 2018]{DBLP:journals/corr/abs-1810-09305}
Koupaee, M. and Wang, W.~Y. (2018).
\newblock Wikihow: {A} large scale text summarization dataset.
\newblock {\em CoRR}, abs/1810.09305.

\bibitem[Ladhak et~al., 2020]{ladhak-etal-2020-wikilingua}
Ladhak, F., Durmus, E., Cardie, C., and McKeown, K. (2020).
\newblock {W}iki{L}ingua: A new benchmark dataset for cross-lingual abstractive
  summarization.
\newblock In {\em Findings of the Association for Computational Linguistics:
  EMNLP 2020}, pages 4034--4048, Online. Association for Computational
  Linguistics.

\bibitem[Lau et~al., 2009]{lau2009interpreting}
Lau, T.~A., Drews, C., and Nichols, J. (2009).
\newblock Interpreting written how-to instructions.
\newblock In {\em IJCAI}, pages 1433--1438.

\bibitem[Li et~al., 2012]{li2012crowdsourcing}
Li, B., Lee-Urban, S., Appling, D.~S., and Riedl, M.~O. (2012).
\newblock Crowdsourcing narrative intelligence.
\newblock {\em Advances in Cognitive systems}, 2(1).

\bibitem[Lin et~al., 2020]{lin-etal-2020-recipe}
Lin, A., Rao, S., Celikyilmaz, A., Nouri, E., Brockett, C., Dey, D., and Dolan,
  B. (2020).
\newblock A recipe for creating multimodal aligned datasets for sequential
  tasks.
\newblock In {\em Proceedings of the 58th Annual Meeting of the Association for
  Computational Linguistics}, pages 4871--4884, Online. Association for
  Computational Linguistics.

\bibitem[Liu et~al., 2019]{liu2019roberta}
Liu, Y., Ott, M., Goyal, N., Du, J., Joshi, M., Chen, D., Levy, O., Lewis, M.,
  Zettlemoyer, L., and Stoyanov, V. (2019).
\newblock Roberta: A robustly optimized bert pretraining approach.
\newblock {\em arXiv preprint arXiv:1907.11692}.

\bibitem[Long et~al., 2016]{long-etal-2016-simpler}
Long, R., Pasupat, P., and Liang, P. (2016).
\newblock Simpler context-dependent logical forms via model projections.
\newblock In {\em Proceedings of the 54th Annual Meeting of the Association for
  Computational Linguistics (Volume 1: Long Papers)}, pages 1456--1465, Berlin,
  Germany. Association for Computational Linguistics.

\bibitem[Lyu et~al., 2021]{lyu-etal-2021-goal}
Lyu, Q., Zhang, L., and Callison-Burch, C. (2021).
\newblock Goal-oriented script construction.
\newblock In {\em Proceedings of the 14th International Conference on Natural
  Language Generation}, pages 184--200, Aberdeen, Scotland, UK. Association for
  Computational Linguistics.

\bibitem[Ma et~al., 2021]{ma-etal-2021-eventplus}
Ma, M.~D., Sun, J., Yang, M., Huang, K.-H., Wen, N., Singh, S., Han, R., and
  Peng, N. (2021).
\newblock {E}vent{P}lus: A temporal event understanding pipeline.
\newblock In {\em Proceedings of the 2021 Conference of the North American
  Chapter of the Association for Computational Linguistics: Human Language
  Technologies: Demonstrations}, pages 56--65, Online. Association for
  Computational Linguistics.

\bibitem[MacMahon et~al., 2006]{macmahon2006walk}
MacMahon, M., Stankiewicz, B., and Kuipers, B. (2006).
\newblock Walk the talk: Connecting language, knowledge, and action in route
  instructions.
\newblock {\em Def}, 2(6):4.

\bibitem[Maeta et~al., 2015]{maeta-etal-2015-framework}
Maeta, H., Sasada, T., and Mori, S. (2015).
\newblock A framework for procedural text understanding.
\newblock In {\em Proceedings of the 14th International Conference on Parsing
  Technologies}, pages 50--60, Bilbao, Spain. Association for Computational
  Linguistics.

\bibitem[Mellish and Evans, 1989]{mellish1989natural}
Mellish, C. and Evans, R. (1989).
\newblock Natural language generation from plans.
\newblock {\em Computational Linguistics}, 15(4):233--249.

\bibitem[Miech et~al., 2019]{miech2019howto100m}
Miech, A., Zhukov, D., Alayrac, J.-B., Tapaswi, M., Laptev, I., and Sivic, J.
  (2019).
\newblock Howto100m: Learning a text-video embedding by watching hundred
  million narrated video clips.
\newblock In {\em Proceedings of the IEEE/CVF International Conference on
  Computer Vision}, pages 2630--2640.

\bibitem[Miller, 1976]{miller1976natural}
Miller, L. (1976).
\newblock Natural language procedures: Guides for programming language design.
\newblock In {\em International Ergonomics Association Meeting, University of
  Maryland}, volume~1, page~76.

\bibitem[Momouchi, 1980]{momouchi-1980-control}
Momouchi, Y. (1980).
\newblock Control structures for actions in procedural texts and {PT}-chart.
\newblock In {\em {COLING} 1980 Volume 1: The 8th International Conference on
  Computational Linguistics}.

\bibitem[Mori et~al., 2014]{mori-etal-2014-flow}
Mori, S., Maeta, H., Yamakata, Y., and Sasada, T. (2014).
\newblock Flow graph corpus from recipe texts.
\newblock In {\em Proceedings of the Ninth International Conference on Language
  Resources and Evaluation ({LREC}'14)}, pages 2370--2377, Reykjavik, Iceland.
  European Language Resources Association (ELRA).

\bibitem[Mujtaba and Mahapatra, 2019]{9070972}
Mujtaba, D. and Mahapatra, N. (2019).
\newblock Recent trends in natural language understanding for procedural
  knowledge.
\newblock In {\em 2019 International Conference on Computational Science and
  Computational Intelligence (CSCI)}, pages 420--424.

\bibitem[Mysore et~al., 2019]{mysore-etal-2019-materials}
Mysore, S., Jensen, Z., Kim, E., Huang, K., Chang, H.-S., Strubell, E.,
  Flanigan, J., McCallum, A., and Olivetti, E. (2019).
\newblock The materials science procedural text corpus: Annotating materials
  synthesis procedures with shallow semantic structures.
\newblock In {\em Proceedings of the 13th Linguistic Annotation Workshop},
  pages 56--64, Florence, Italy. Association for Computational Linguistics.

\bibitem[Nguyen et~al., 2017]{nguyen-etal-2017-sequence}
Nguyen, D.~Q., Nguyen, D.~Q., Chu, C.~X., Thater, S., and Pinkal, M. (2017).
\newblock Sequence to sequence learning for event prediction.
\newblock In {\em Proceedings of the Eighth International Joint Conference on
  Natural Language Processing (Volume 2: Short Papers)}, pages 37--42, Taipei,
  Taiwan. Asian Federation of Natural Language Processing.

\bibitem[Nighojkar and Licato, 2021]{nighojkar-licato-2021-improving}
Nighojkar, A. and Licato, J. (2021).
\newblock Improving paraphrase detection with the adversarial paraphrasing
  task.
\newblock In {\em Proceedings of the 59th Annual Meeting of the Association for
  Computational Linguistics and the 11th International Joint Conference on
  Natural Language Processing (Volume 1: Long Papers)}, pages 7106--7116,
  Online. Association for Computational Linguistics.

\bibitem[Pareti, 2018]{pareti2018representation}
Pareti, P. (2018).
\newblock Representation and execution of human know-how on the web.

\bibitem[Pareti et~al., 2014a]{10.1145/2567948.2578846}
Pareti, P., Klein, E., and Barker, A. (2014a).
\newblock A semantic web of know-how: Linked data for community-centric tasks.
\newblock In {\em Proceedings of the 23rd International Conference on World
  Wide Web}, WWW '14 Companion, page 1011–1016, New York, NY, USA.
  Association for Computing Machinery.

\bibitem[Pareti et~al., 2014b]{pareti2014integrating}
Pareti, P., Testu, B., Ichise, R., Klein, E., and Barker, A. (2014b).
\newblock Integrating know-how into the linked data cloud.
\newblock In {\em International Conference on Knowledge Engineering and
  Knowledge Management}, pages 385--396. Springer.

\bibitem[Paris et~al., 2005]{paris_automatically_2005}
Paris, C., Colineau, N., Shijian, L., and Linden, K.~V. (2005).
\newblock Automatically generating effective online help.

\bibitem[Paris et~al., 2002]{10.1145/584955.584977}
Paris, C., Linden, K.~V., and Lu, S. (2002).
\newblock Automated knowledge acquisition for instructional text generation.
\newblock In {\em Proceedings of the 20th Annual International Conference on
  Computer Documentation}, SIGDOC '02, page 142–151, New York, NY, USA.
  Association for Computing Machinery.

\bibitem[Paris et~al., 1995]{paris1995support}
Paris, C., Vander~Linden, K., Fischer, M., Hartley, A., Pemberton, L., Power,
  R., and Scott, D. (1995).
\newblock A support tool for writing multilingual instructions.
\newblock In {\em International Joint Conference on Artificial Intelligence},
  volume~14, pages 1398--1404. LAWRENCE ERLBAUM ASSOCIATES LTD.

\bibitem[Park and Motahari~Nezhad, 2018]{park2018learning}
Park, H. and Motahari~Nezhad, H.~R. (2018).
\newblock Learning procedures from text: Codifying how-to procedures in deep
  neural networks.
\newblock In {\em Companion Proceedings of the The Web Conference 2018}, pages
  351--358.

\bibitem[Perkowitz et~al., 2004]{10.1145/988672.988750}
Perkowitz, M., Philipose, M., Fishkin, K., and Patterson, D.~J. (2004).
\newblock Mining models of human activities from the web.
\newblock In {\em Proceedings of the 13th International Conference on World
  Wide Web}, WWW '04, page 573–582, New York, NY, USA. Association for
  Computing Machinery.

\bibitem[Puig et~al., 2018]{puig2018virtualhome}
Puig, X., Ra, K., Boben, M., Li, J., Wang, T., Fidler, S., and Torralba, A.
  (2018).
\newblock Virtualhome: Simulating household activities via programs.
\newblock In {\em Proceedings of the IEEE Conference on Computer Vision and
  Pattern Recognition}, pages 8494--8502.

\bibitem[Raffel et~al., 2020]{2020t5}
Raffel, C., Shazeer, N., Roberts, A., Lee, K., Narang, S., Matena, M., Zhou,
  Y., Li, W., and Liu, P.~J. (2020).
\newblock Exploring the limits of transfer learning with a unified text-to-text
  transformer.
\newblock {\em Journal of Machine Learning Research}, 21(140):1--67.

\bibitem[Rajagopal et~al., 2020]{rajagopal-etal-2020-ask}
Rajagopal, D., Tandon, N., Clark, P., Dalvi, B., and Hovy, E. (2020).
\newblock What-if {I} ask you to explain: Explaining the effects of
  perturbations in procedural text.
\newblock In {\em Findings of the Association for Computational Linguistics:
  EMNLP 2020}, pages 3345--3355, Online. Association for Computational
  Linguistics.

\bibitem[Regneri et~al., 2010]{regneri-etal-2010-learning}
Regneri, M., Koller, A., and Pinkal, M. (2010).
\newblock Learning script knowledge with web experiments.
\newblock In {\em Proceedings of the 48th Annual Meeting of the Association for
  Computational Linguistics}, pages 979--988, Uppsala, Sweden. Association for
  Computational Linguistics.

\bibitem[Reimers and Gurevych, 2019]{reimers-gurevych-2019-sentence}
Reimers, N. and Gurevych, I. (2019).
\newblock Sentence-{BERT}: Sentence embeddings using {S}iamese {BERT}-networks.
\newblock In {\em Proceedings of the 2019 Conference on Empirical Methods in
  Natural Language Processing and the 9th International Joint Conference on
  Natural Language Processing (EMNLP-IJCNLP)}, pages 3982--3992, Hong Kong,
  China. Association for Computational Linguistics.

\bibitem[Sakaguchi et~al., 2021]{sakaguchi-etal-2021-proscript-partially}
Sakaguchi, K., Bhagavatula, C., Le~Bras, R., Tandon, N., Clark, P., and Choi,
  Y. (2021).
\newblock pro{S}cript: Partially ordered scripts generation.
\newblock In {\em Findings of the Association for Computational Linguistics:
  EMNLP 2021}, pages 2138--2149, Punta Cana, Dominican Republic. Association
  for Computational Linguistics.

\bibitem[Schank, 1977]{schank_scripts_1977}
Schank, R.~C. (1977).
\newblock {\em Scripts, plans, goals, and understanding : an inquiry into human
  knowledge structures /}.
\newblock L. Erlbaum Associates ;, Hillsdale, N.J. :.

\bibitem[Shapiro and Iw{\'a}nska, 2000]{shapiro2000natural}
Shapiro, S.~C. and Iw{\'a}nska, L.~M. (2000).
\newblock {\em Natural language processing and knowledge representation:
  language for knowledge and knowledge for language}.
\newblock Citeseer.

\bibitem[Takechi et~al., 2003]{takechi-etal-2003-feature}
Takechi, M., Tokunaga, T., Matsumoto, Y., and Tanaka, H. (2003).
\newblock Feature selection in categorizing procedural expressions.
\newblock In {\em Proceedings of the Sixth International Workshop on
  Information Retrieval with {A}sian Languages}, pages 49--56, Sapporo, Japan.
  Association for Computational Linguistics.

\bibitem[Tandon et~al., 2019]{tandon-etal-2019-wiqa}
Tandon, N., Dalvi, B., Sakaguchi, K., Clark, P., and Bosselut, A. (2019).
\newblock {WIQA}: A dataset for {``}what if...{''} reasoning over procedural
  text.
\newblock In {\em Proceedings of the 2019 Conference on Empirical Methods in
  Natural Language Processing and the 9th International Joint Conference on
  Natural Language Processing (EMNLP-IJCNLP)}, pages 6076--6085, Hong Kong,
  China. Association for Computational Linguistics.

\bibitem[Tandon et~al., 2020]{tandon-etal-2020-dataset}
Tandon, N., Sakaguchi, K., Dalvi, B., Rajagopal, D., Clark, P., Guerquin, M.,
  Richardson, K., and Hovy, E. (2020).
\newblock A dataset for tracking entities in open domain procedural text.
\newblock In {\em Proceedings of the 2020 Conference on Empirical Methods in
  Natural Language Processing (EMNLP)}, pages 6408--6417, Online. Association
  for Computational Linguistics.

\bibitem[Tasse and Smith, 2008]{tasse2008sour}
Tasse, D. and Smith, N.~A. (2008).
\newblock Sour cream: Toward semantic processing of recipes.
\newblock {\em Carnegie Mellon University, Pittsburgh, Tech. Rep.
  CMU-LTI-08-005}.

\bibitem[Vashishtha et~al., 2020]{vashishtha-etal-2020-temporal}
Vashishtha, S., Poliak, A., Lal, Y.~K., Van~Durme, B., and White, A.~S. (2020).
\newblock Temporal reasoning in natural language inference.
\newblock In {\em Findings of the Association for Computational Linguistics:
  EMNLP 2020}, pages 4070--4078, Online. Association for Computational
  Linguistics.

\bibitem[Wahlster et~al., 1993]{wahlster1993plan}
Wahlster, W., Andr{\'e}, E., Finkler, W., Profitlich, H.-J., and Rist, T.
  (1993).
\newblock Plan-based integration of natural language and graphics generation.
\newblock {\em Artificial intelligence}, 63(1-2):387--427.

\bibitem[Wanzare et~al., 2016]{wanzare-etal-2016-crowdsourced}
Wanzare, L. D.~A., Zarcone, A., Thater, S., and Pinkal, M. (2016).
\newblock A crowdsourced database of event sequence descriptions for the
  acquisition of high-quality script knowledge.
\newblock In {\em Proceedings of the Tenth International Conference on Language
  Resources and Evaluation ({LREC}'16)}, pages 3494--3501, Portoro{\v{z}},
  Slovenia. European Language Resources Association (ELRA).

\bibitem[Yamakata et~al., 2016]{7574705}
Yamakata, Y., Imahori, S., Maeta, H., and Mori, S. (2016).
\newblock A method for extracting major workflow composed of ingredients,
  tools, and actions from cooking procedural text.
\newblock In {\em 2016 IEEE International Conference on Multimedia Expo
  Workshops (ICMEW)}, pages 1--6.

\bibitem[Yang et~al., 2021]{yang-etal-2021-visual}
Yang, Y., Panagopoulou, A., Lyu, Q., Zhang, L., Yatskar, M., and
  Callison-Burch, C. (2021).
\newblock Visual goal-step inference using wiki{H}ow.
\newblock In {\em Proceedings of the 2021 Conference on Empirical Methods in
  Natural Language Processing}, pages 2167--2179, Online and Punta Cana,
  Dominican Republic. Association for Computational Linguistics.

\bibitem[Zellers et~al., 2019]{zellers-etal-2019-hellaswag}
Zellers, R., Holtzman, A., Bisk, Y., Farhadi, A., and Choi, Y. (2019).
\newblock {H}ella{S}wag: Can a machine really finish your sentence?
\newblock In {\em Proceedings of the 57th Annual Meeting of the Association for
  Computational Linguistics}, pages 4791--4800, Florence, Italy. Association
  for Computational Linguistics.

\bibitem[Zhang et~al., 2020a]{zhang-etal-2020-analogous}
Zhang, H., Chen, M., Wang, H., Song, Y., and Roth, D. (2020a).
\newblock Analogous process structure induction for sub-event sequence
  prediction.
\newblock In {\em Proceedings of the 2020 Conference on Empirical Methods in
  Natural Language Processing (EMNLP)}, pages 1541--1550, Online. Association
  for Computational Linguistics.

\bibitem[Zhang et~al., 2020b]{zhang-etal-2020-reasoning}
Zhang, L., Lyu, Q., and Callison-Burch, C. (2020b).
\newblock Reasoning about goals, steps, and temporal ordering with {W}iki{H}ow.
\newblock In {\em Proceedings of the 2020 Conference on Empirical Methods in
  Natural Language Processing (EMNLP)}, pages 4630--4639, Online. Association
  for Computational Linguistics.

\bibitem[Zhang et~al., 2019]{zhang2019bertscore}
Zhang, T., Kishore, V., Wu, F., Weinberger, K.~Q., and Artzi, Y. (2019).
\newblock Bertscore: Evaluating text generation with bert.
\newblock {\em arXiv preprint arXiv:1904.09675}.

\bibitem[Zhang et~al., 2012]{zhang-etal-2012-automatically}
Zhang, Z., Webster, P., Uren, V., Varga, A., and Ciravegna, F. (2012).
\newblock Automatically extracting procedural knowledge from instructional
  texts using natural language processing.
\newblock In {\em Proceedings of the Eighth International Conference on
  Language Resources and Evaluation ({LREC}'12)}, pages 520--527, Istanbul,
  Turkey. European Language Resources Association (ELRA).

\bibitem[Zhou et~al., 2019a]{zhou-etal-2019-going}
Zhou, B., Khashabi, D., Ning, Q., and Roth, D. (2019a).
\newblock {``}going on a vacation{''} takes longer than {``}going for a
  walk{''}: A study of temporal commonsense understanding.
\newblock In {\em Proceedings of the 2019 Conference on Empirical Methods in
  Natural Language Processing and the 9th International Joint Conference on
  Natural Language Processing (EMNLP-IJCNLP)}, pages 3363--3369, Hong Kong,
  China. Association for Computational Linguistics.

\bibitem[Zhou et~al., 2018]{zhou2018towards}
Zhou, L., Xu, C., and Corso, J.~J. (2018).
\newblock Towards automatic learning of procedures from web instructional
  videos.
\newblock In {\em Thirty-Second AAAI Conference on Artificial Intelligence}.

\bibitem[Zhou et~al., 2022]{zhou-etal-2022-show}
Zhou, S., Zhang, L., Yang, Y., Lyu, Q., Yin, P., Callison-Burch, C., and
  Neubig, G. (2022).
\newblock Show me more details: Discovering hierarchies of procedures from
  semi-structured web data.
\newblock In {\em Proceedings of the 60th Annual Meeting of the Association for
  Computational Linguistics (Volume 1: Long Papers)}, Dublin, Ireland.
  Association for Computational Linguistics.

\bibitem[Zhou et~al., 2019b]{zhou-etal-2019-learning-household}
Zhou, Y., Shah, J., and Schockaert, S. (2019b).
\newblock Learning household task knowledge from {W}iki{H}ow descriptions.
\newblock In {\em Proceedings of the 5th Workshop on Semantic Deep Learning
  (SemDeep-5)}, pages 50--56, Macau, China. Association for Computational
  Linguistics.

\end{thebibliography}
\end{document}